\newcommand{\lora}{{LoRA}}
\newcommand{\ourmethodslim}{{LoRASculpt}}

\newcommand{\zeroshot}{{Zero-shot}}
\newcommand{\dora}{{DoRA}}
\newcommand{\ltworeg}{{L2-Reg}}
\newcommand{\orthreg}{{Orth-Reg}}
\newcommand{\tailor}{{Model Tailor}}
\newcommand{\dare}{{DARE}}

\newcommand{\iconqa}{{IconQA}}
\newcommand{\coco}{{COCO-Caption}}
\newcommand{\okvqa}{{OKVQA}}
\newcommand{\ocrvqa}{{OCRVQA}}
\newcommand{\gqa}{{GQA}}
\newcommand{\textvqa}{{TextVQA}}

\newcommand{\redup}[1]{$_{\color{RedOrange}\uparrow #1}$}

\documentclass[10pt,twocolumn,letterpaper]{article}

\usepackage{cvpr}

\definecolor{cvprblue}{rgb}{0.21,0.49,0.74}
\usepackage[pagebackref,breaklinks,colorlinks,allcolors=cvprblue]{hyperref}

\usepackage{amsmath}
\usepackage{amsthm}
\usepackage{cases}
\usepackage{algorithmic}
\usepackage{multirow}

\usepackage{xcolor}
\definecolor{myRed}{HTML}{FF6F61} 
\definecolor{myGreen}{HTML}{2B7A78} 

\theoremstyle{plain} 
\newtheorem{theorem}{Theorem}[section]

\usepackage[ruled]{algorithm2e}
\usepackage[export]{adjustbox}
\usepackage{threeparttable}
\usepackage{subcaption}
\usepackage{multirow}
\usepackage{enumitem}
\usepackage{newfloat}
\usepackage{listings}
\usepackage{colortbl}
\usepackage{arydshln}
\usepackage{amsfonts}
\usepackage{caption}
\usepackage{wrapfig}
\usepackage{amssymb}
\usepackage{pifont}
\usepackage{float}
\usepackage{bbm}
\usepackage{bm}
\usepackage{mathrsfs}  

\definecolor{mygray}{gray}{.9}
\definecolor{mygreen}{RGB}{93,173,85}
\definecolor{mywarning}{RGB}{233,144,61}

\definecolor{DarkBlue}{RGB}{64,101,149}
\definecolor{azure}{rgb}{0.0, 0.5, 1.0}
\definecolor{gray}{rgb}{0.3, 0.3, 0.3}
\definecolor{DarkGreen}{RGB}{42,110,63}
\definecolor{DarkYellow}{RGB}{191,144,0}
\definecolor{DarkRed}{rgb}{0.6, 0, 0}

\newcolumntype{x}[1]{>{\centering\arraybackslash}p{#1pt}}
\newcolumntype{I}{!{\vrule width 1pt}}

\usepackage{tcolorbox}
\tcbuselibrary{theorems}
\definecolor{lightgray}{gray}{.9}
\definecolor{deepgray}{gray}{.8}
\tcbset{highlight math/.append style={left=0mm,right=0mm,top=0mm,bottom=0mm, colframe=white}}

\makeatletter
\newcommand{\thickhline}{%
    \noalign {\ifnum 0=`}\fi \hrule height 1pt
    \futurelet \reserved@a \@xhline
}



\usepackage[capitalize]{cleveref}
\crefname{proposition}{Prop.}{Props.}
\crefname{section}{Sec.}{Secs.}
\crefname{table}{Tab.}{Tabs.}
\newcommand{\pub}[1]{{\color{gray}{\footnotesize{[{#1}]}}}}

\usepackage{xspace}
\makeatletter
\DeclareRobustCommand\onedot{\futurelet\@let@token\@onedot}
\def\@onedot{\ifx\@let@token.\else.\null\fi\xspace}

\definecolor{darksalmon}{rgb}{0.91, 0.59, 0.48}
\definecolor{emerald}{rgb}{0.31, 0.78, 0.47}
\definecolor{green(pigment)}{rgb}{0.0, 0.65, 0.31}
\definecolor{amaranth}{rgb}{0.9, 0.17, 0.31}
\definecolor{iris}{rgb}{0.35, 0.31, 0.81}
\definecolor{uu}{rgb}{0.95, 0.51, 0.51}
\definecolor{spirodiscoball}{rgb}{0.06, 0.75, 0.99}
\usepackage{bbding}

\usepackage{xspace}
\newcommand{\ourmethod}{{\fontfamily{lmtt}\selectfont \textbf{LoRASculpt}}\xspace}

\definecolor{P1}{RGB}{241, 18, 12}
\definecolor{P2}{RGB}{49, 71, 129}
\definecolor{myblue}{RGB}{19, 10, 114}
\usepackage{mdframed}
\usepackage{xcolor}

\usepackage[pagebackref,breaklinks,colorlinks,allcolors=cvprblue]{hyperref}

\def\confName{CVPR}
\def\confYear{2025}

\title{\includegraphics[width=0.04\textwidth]{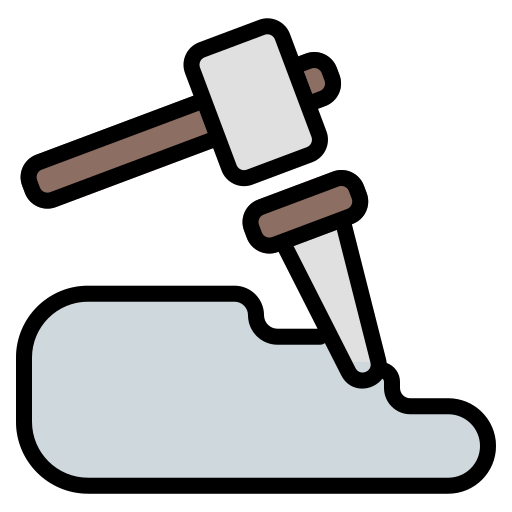}  LoRASculpt: Sculpting LoRA for Harmonizing General and Specialized Knowledge in Multimodal Large Language Models}

\author{
Jian Liang\footnotemark[1], \hspace{2pt} Wenke Huang\thanks{Equal contributions}, \hspace{2pt} Guancheng Wan\footnotemark[1], \hspace{2pt} Qu Yang, \hspace{2pt} Mang Ye\thanks{Corresponding Author}
\\ National Engineering Research Center for Multimedia Software, \\
School of Computer Science, Wuhan University.\\
\texttt{\small{\{jianliang, wenkehuang, yemang\}@whu.edu.cn}}
}

\begin{document}
\maketitle
\begin{abstract}

While Multimodal Large Language Models (MLLMs) excel at generalizing across modalities and tasks, effectively adapting them to specific downstream tasks while simultaneously retaining both general and specialized knowledge remains challenging. 
Although Low-Rank Adaptation (LoRA) is widely used to efficiently acquire specialized knowledge in MLLMs, it introduces substantial harmful redundancy during visual instruction tuning, which exacerbates the forgetting of general knowledge and degrades downstream task performance.
To address this issue, we propose \ourmethod{} to eliminate harmful redundant parameters, thereby harmonizing general and specialized knowledge.
Specifically, under theoretical guarantees, we introduce sparse updates into LoRA to discard redundant parameters effectively. Furthermore, we propose a Conflict Mitigation Regularizer to refine the update trajectory of LoRA, mitigating knowledge conflicts with the pretrained weights.
Extensive experimental results demonstrate that even at very high degree of sparsity ($\le$ 5\%), our method simultaneously enhances generalization and downstream task performance. This confirms that our approach effectively mitigates the catastrophic forgetting issue and further promotes knowledge harmonization in MLLMs.

\end{abstract}    
\section{Introduction}
\label{sec:intro}

\begin{figure}[t]
\centering
\includegraphics[width=0.92\linewidth]{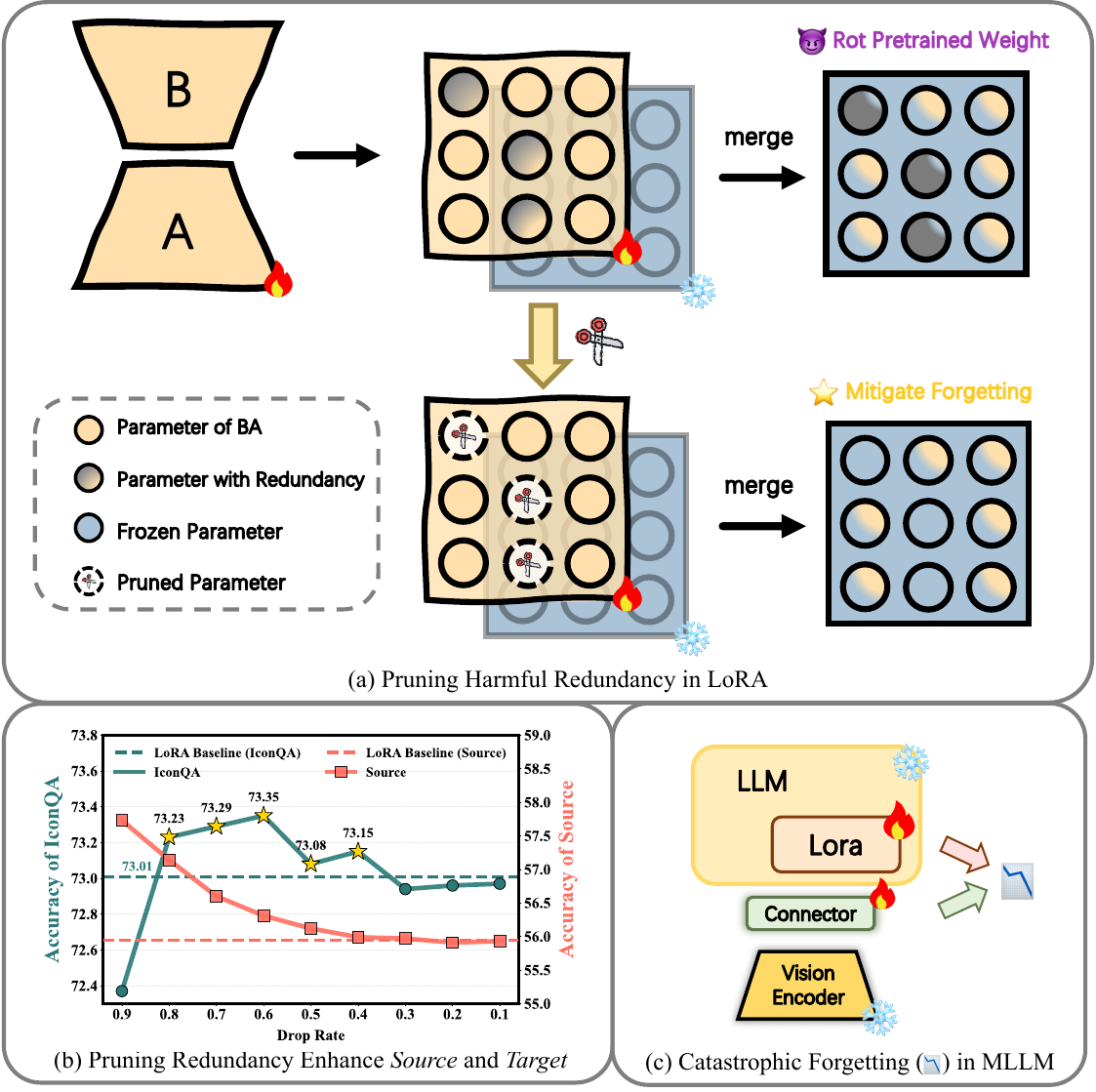}
\vspace{-5pt}
\captionsetup{font=small}
\caption{\textbf{Motivation}. Fine-tuning MLLM with LoRA on downstream tasks generates numerous harmful redundancy. (a) 
Illustration of post-pruning LoRA to reduce redundancy. (b) Simply pruning harmful redundancy in LoRA based on magnitude \textcolor{myRed}{\textit{\textbf{reduces forgetting of pretrained knowledge}} (\textit{Source})}, and even \textcolor{myGreen}{\textit{\textbf{enhances downstream task performance}} (\textit{Target})}. (c) Current MLLMs suffer from catastrophic forgetting in both LLM and connector with fewer parameters, as detailed in \cref{sec:cf_connector}.
}
\label{fig:motivation}
\vspace{-15pt}
\end{figure}

Multimodal large language models (MLLMs) are known for their strong generalization abilities, making them adaptable across modalities and tasks \cite{LLaVA15_CVPR24,chen2024internvl,li2024monkey,wang2024qwen2,li2024llava}.
Visual instruction tuning \cite{LLaVA_NeurIPS23} is often applied to enhance performance on specific downstream tasks. However, as the model size grows, full fine-tuning becomes resource-intensive. 
Consequently, \lora{}-based methods \cite{LoRA_ICLR22,DoRA_ICML24,kopiczko2024vera,AdaLoRA_arXiv23} have gained popularity for their efficiency, preserving the model architecture without additional inference costs. 
While \lora{} induces less forgetting than full fine-tuning \cite{biderman2024lora,ni2023forgetting}, merging it with the pretrained model still modifies all parameters, posing a risk of severe generalization loss.
The phenomenon of catastrophic forgetting in LoRA has been widely observed across several studies \cite{dou2024loramoe,yang2024corda,han2024slim,ModelTailor_ICML24,yu2024boosting}.
Furthermore, this problem is especially pronounced in MLLMs due to their complex architectures, cross-modal data differences, and task interference, posing heightened challenges in maintaining general knowledge while improving performance on downstream tasks \cite{sung2023ecoflap,shen2024multimodal,zhai2024investigating,jiang2024effectiveness}.
These issues make it challenging to balance general and specialized knowledge in deploying MLLMs, where both capabilities are essential.

To enable MLLMs to learn downstream tasks while preserving generalization, an intuitive approach is to minimize conflicts with pre-existing general knowledge during new task learning.
However, although reducing conflicts can help preserve generalization, it may compromise performance on downstream tasks, raising the question: \textit{How can we better balance this trade-off?}
We observe a key phenomenon: large foundation models often exhibit \textit{\textbf{significant parameter redundancy}} in delta weights after supervised fine-tuning (SFT) \cite{DARE_ICML24}.
Furthermore, recent research has successfully eliminated redundant parameters in LoRA through post-pruning, showing that LoRA retains harmful redundancy after training, which not only fails to effectively serve the target task but also erodes pretrained knowledge \cite{ModelTailor_ICML24}. Our additional experiments confirm that these harmful redundant parameters in LoRA are abundant and explicit: by simply applying post-pruning to the trained LoRA based on parameter magnitudes, both the general and specialized capabilities of MLLMs can be simultaneously improved, as is shown in \cref{fig:motivation} (a) and (b). However, as sparsity increases, post-pruning methods tend to reduce downstream task performance \cite{zhang2023loraprune}. 
Motivated by the substantial harmful redundancy in LoRA after SFT, we consider \textbf{sculpting LoRA for redundancy reduction} during training, to ensure both generalization and specialization ability.

However, mitigating LoRA redundancy faces two crucial challenges: First, unlike standard parameter modules, sparsity cannot be applied directly to the product matrix BA of LoRA during training, as the pruned BA cannot be decomposed back into B and A. Additionally, directly sparsifying B and A does not necessarily yield a sparse product matrix BA \cite{yuster2005fast}. Thus, this leads to the {\color{Black}\textbf{LoRA Sparsity Dilemma}}, which fails to ensure the necessary sparsity of BA. This raises the first challenge: \hypertarget{Q1} {\textbf{\uppercase\expandafter{\romannumeral1})}} {\textit{How to ensure the necessary sparsity of LoRA for redundancy reduction?}} Second, although removing redundant parameters can restore pretrained general knowledge and thus mitigate forgetting, it does not address a key issue: the remaining task-specific parameters may still risk conflicting with important parameters in the pretrained model. LoRA updates faithfully follow the current empirical loss minimization, thus encountering optimization interference from pretrained knowledge behavior, which brings the {\color{Black}\textbf{LoRA Optimization Conflict}}, leading to general knowledge forgetting. Thus, this raises the second key challenge: \hypertarget{Q2}{\textbf{\uppercase\expandafter{\romannumeral2})}} {\textit{How to calibrate LoRA update trajectory to alleviate knowledge conflicts?}}

In response to these challenges, we introduce a novel framework named \textbf{LoRASculpt}, which integrates sparse updates into LoRA for redundancy reduction, and introduces a Conflict Mitigation Regularizer for knowledge harmonization. 
Specifically, to address challenge \hyperlink{Q1}{\textbf{\uppercase\expandafter{\romannumeral1})}} , we implement \textit{Sparsifying LoRA for Redundancy Reduction}, introducing sparsity into the training process of LoRA with rigorous theoretical guarantees. 
We prove the expected sparsity of the product of two sparse low-rank matrices and derive an upper bound for the probability of exceeding the expected sparsity, thus effectively resolving the LoRA Sparsity Dilemma.
For challenge \hyperlink{Q2}{\textbf{\uppercase\expandafter{\romannumeral2})}} , we implement \textit{Regularizing LoRA for Knowledge Harmonization}. A pretrained knowledge-guided Conflict Mitigation Regularizer is proposed to adjust the optimization trajectory of LoRA, directing task-specific knowledge injection into parameter locations that are less critical in the pretrained model, thereby minimizing the LoRA Optimization Conflict. We further discuss the synergy between the two components in mitigating conflicts between general and specialized knowledge.

Extensive experiments on VQA and Captioning tasks demonstrate that \ourmethod{} effectively mitigates generalization loss in MLLMs. 
Moreover, by appropriately mitigating the optimization conflict and pruning the redundancy of \lora{}, our approach can even enhance performance on downstream tasks. We reveal that severe forgetting also occurs in connector module, which typically has fewer parameters than LLM module. Adapting our method to connectors with simple modifications can further boost performance. 
Our main contributions are summarized as follows:

\begin{itemize}[leftmargin=*]

\item[\ding{182}] \textbf{\textit{Re-examining Forgetting in MLLM Trained with LoRA.}} Our findings indicate that LoRA exhibits abundant harmful redundancy after SFT in MLLMs, which undermines both generalization and specialization performance, even within the connector with fewer parameters.

\item[\ding{183}] \textbf{\textit{Novel Parameter-Efficient Framework for Mitigating Forgetting in MLLMs.}} Building on the phenomenon of parameter redundancy in LoRA, we effectively mitigate forgetting by addressing the LoRA Sparsity Dilemma and LoRA Optimization Conflict.

\item[\ding{184}] \textbf{\textit{Theoretical Guarantees and Experimental Validation.}} We provide theoretical guarantees for introducing sparse updates into LoRA to reduce redundant parameters, and further demonstrate the effectiveness and robustness of our framework through comprehensive experiments.
\end{itemize}

\section{Related Works}
\label{sec:formatting}

\noindent\textbf{Multimodal Large Language Models.} With the advancement of multimodal learning, the generalization of small-scale multimodal models has been widely explored \cite{FCCLPlus_TPAMI23,hu2024fedcross,FLSurveyandBenchmarkforGenRobFair_TPAMI24}, and recently extended to Multimodal Large Language Models (MLLMs) \cite{QwenVL_arXiv23,LLaVA_NeurIPS23,LLaVA15_CVPR24,huang2024empirical,ye2025survey,bai2025chat}. MLLMs have significantly improved the multimodal understanding, cross-modal reasoning, and problem-solving abilities by integrating visual encoders \cite{CLIP_ICML21,zhai2023sigmoid} with LLMs \cite{LLaMA_arXiv23,Vicuna_arXiv23,touvron2023llama,dubey2024llama}, demonstrating strong generalization ability. The connector module such as MLPs and the Q-Former \cite{BLIPv2_ICML23} enhances alignment between visual and language features, facilitating smooth and efficient interaction between the visual encoder and language model. Visual instruction tuning \cite{LLaVA_NeurIPS23} further adapts MLLMs for specific downstream tasks, improving task-specific performance \cite{ModelTailor_ICML24,huang2024learn}.

\noindent\textbf{Catastrophic Forgetting.}
Deep learning models often suffer from catastrophic forgetting, where previously learned knowledge is lost when learning new tasks \cite{wang2024comprehensive,mcclelland1995there,mccloskey1989catastrophic,wang2022learning,li2024unleashing,FCCL_CVPR22,FCCLPlus_TPAMI23}. Various continual learning algorithms have been proposed to address this issue, generally categorized into rehearsal-based \cite{chaudhry2019tiny,lopez2017gradient,riemer2018learning}, regularization-based \cite{kirkpatrick2017overcoming,schwarz2018progress,liu2018rotate}, and architecture-based \cite{yoon2017lifelong,wang2023rehearsal,razdaibiedina2023progressive,zhong2024moextend} approaches. Some traditional methods have explored using pruning to mitigate forgetting \cite{wang2022sparcl,sokar2021spacenet,wang2020learn,mallya2018packnet,mallya2018piggyback}; however, these are mostly designed for small models and rely heavily on full-model fine-tuning, making them less applicable for fine-tuning large foundation models. 
In the era of large foundation models, the traditional forgetting problem shifts to maintaining the model’s strong generalization ability after downstream task fine-tuning \cite{dou2024loramoe,han2024slim,ModelTailor_ICML24}. Additionally, these models are often closed-source, with only model weight accessible.
Recently, SPU \cite{zhang2024overcoming} demonstrates the feasibility of sparse updates for fine-tuning CLIP to mitigate general knowledge forgetting. 
Other studies have explored the forgetting issue when training LLMs with LoRA. LoRAMoE \cite{dou2024loramoe} and SLIM \cite{han2024slim} apply MoE architectures within LoRA to address generalization loss in LLMs. CorDA \cite{yang2024corda} builds task-aware adapters through context-guided weight decomposition to preserve world knowledge. 
Model Tailor \cite{ModelTailor_ICML24} pioneers a parameter-efficient solution to MLLM forgetting, but this post-training approach carries the risk of leading to suboptimal downstream performance \cite{zhang2023loraprune}.
\section{Methodology}

\subsection{Preliminary}
\label{sec:Preliminary}

\noindent\textbf{Low-Rank Adaptation} (LoRA) \cite{LoRA_ICLR22} is a parameter-efficient fine-tuning method for adapting large foundation models with minimal computational cost. Instead of updating all parameters, LoRA learns weight changes by introducing low-rank matrices \( B \) and \( A \), so that fine-tuned weights \( W \) are expressed as:
\begin{equation}
\setlength\abovedisplayskip{3pt} \setlength\belowdisplayskip{3pt}
W = W_0 + \Delta W = W_0 + BA,
\end{equation}
where \( W_0 \) is the pre-trained weight matrix, and \( r \ll \min(p, q) \) denotes the low rank of LoRA.

\vspace{0.5em}
\noindent\textbf{Harmful Redundant Delta Weights after SFT.}
Recent study shows that the delta vectors after supervised fine-tuning (SFT) in LLMs are highly redundant. Removing most of these parameters has minimal impact on performance \cite{yu2024language}.
Model Tailor reached similar conclusions on MLLMs, suggesting that redundant parameters in LoRA may further degrade downstream task performance \cite{ModelTailor_ICML24}. Our further experiments show that harmful redundancy in LoRA are abundant and explicit. Simply pruning low-magnitude parameters can mitigate generalization loss and even improve downstream task performance, as shown in \cref{fig:motivation}. We restate this key phenomenon as follows:
\definecolor{cadetblue}{RGB}{95,158,160} 
\definecolor{keywordcolor}{RGB}{178,34,34} 
\begin{mdframed}[backgroundcolor=cadetblue!10, linewidth=0.8pt, linecolor=cadetblue!80, roundcorner=5pt]
\textbf{\textcolor{keywordcolor}{Harmful Parameter Redundancy in LoRA}}:
\textit{LoRA introduces substantial and explicit \textbf{harmful redundancy} during SFT, which not only diminishes the generalization ability of MLLM but can also fail to contribute positively on downstream tasks. }
\end{mdframed}

\subsection{Proposed Method}
\label{sec:Proposed_Method}
Inspired by our observations in \cref{sec:Preliminary} that LoRA generates substantial harmful redundancy when SFT on MLLMs, we propose \ourmethod{} to finely sculpt LoRA, enabling the elimination of redundant parameters and achieving a balance between general and specialized knowledge. We divide our method into two main components: establishing the feasibility of sparsifying LoRA for redundancy reduction (\cref{sec:Sparsifying_LoRA_for_Redundancy_Reduction}), and regularizing LoRA for knowledge harmonization (\cref{sec:Regularizing_LoRA_for_Knowledge_Harmonization}).
Finally, we present the adaptation of LoRASculpt for the MLLM connector.

\subsubsection{Sparsifying LoRA for Redundancy Reduction}
\label{sec:Sparsifying_LoRA_for_Redundancy_Reduction}
Given the presence of substantial harmful redundancy in LoRA, we aim to introduce sparsity by pruning redundant parameters. In this section, we demonstrate the feasibility of incorporating sparse updates into LoRA training, effectively compressing downstream task knowledge into a critical subset of sparse parameters.

However, as mentioned in challenge \hyperlink{Q1}{\textbf{\uppercase\expandafter{\romannumeral1})}}, introducing sparsity in LoRA leads to the \textit{\textbf{LoRA Sparsity Dilemma}}—incorporating sparsity into LoRA is not trivial. We claim that, although the product of two sparse matrices is not necessarily sparse, under the conditions of low-rank matrices in LoRA and high sparsity, this sparsity can be theoretically bounded with formal guarantees.

To implement pruning, we need to evaluate parameter importance, a considerable amount of research has been conducted, which can be broadly categorized into magnitude-based \cite{LTH_ICLR19,Wanda_ICLR24,MagnitudePurning_NeurIPS15} and gradient-based \cite{TaSL_ACL24,wang2022sparcl,StableCL_NeurIPS20} approaches. Although gradient-based methods often combine magnitude and gradient information to improve importance estimation, when using LoRA in resource-constrained scenarios, minimizing memory usage and computational cost during fine-tuning becomes critical. Therefore, we adopt the simple yet effective and efficient importance evaluation method that leverages the magnitude of parameters as a proxy for importance. 
Specifically, after a warm-up period, we assess the magnitude of each parameter and prune the less important parameters within the low-rank matrices $A$ and $B$ in LoRA, resulting in pruned matrices \( \widetilde{A} \) and \( \widetilde{B} \) as follows:
\begin{equation}
\setlength\abovedisplayskip{3pt} \setlength\belowdisplayskip{3pt}
\left\{
\begin{array}{l}
\widetilde{A} = M_A \odot A \\
\widetilde{B} = M_B \odot B  ,
\end{array}
\right.
\label{eq:sparsify_AB}
\end{equation}
where \( M_A \) and \( M_B \) are sparsity masks defined by the sparsity levels \( s_A \) and \( s_B \), respectively. Specifically, \( M_A \) and \( M_B \) are constructed as follows:
\begin{equation}
\setlength\abovedisplayskip{3pt} \setlength\belowdisplayskip{3pt}
M_A = \operatorname{Mask}(A, s_A), \quad M_B = \operatorname{Mask}(B, s_B),
\label{eq:sparsify}
\end{equation}
where \( \operatorname{Mask}(X, s) \) denotes a function that generates a binary mask for matrix \( X \), retaining a fraction \( s \) of elements based on importance (i.e., magnitude).
Unlike the magnitude of overall model, the magnitude of LoRA reflects accumulated gradient updates on downstream tasks, i.e., delta parameters \cite{DARE_ICML24}. This cumulative effect makes LoRA more robust than single-step gradient estimation, as it smooths out variations across steps and reduces noise sensitivity.

However, this introduces a potential issue: \textit{\textbf{the product of two sparse matrices \bm{$B$} and \bm{$A$}, denoted as \bm{$BA$}, is not necessarily sparse \cite{yuster2005fast}.}} If $BA$ cannot maintain sparsity, then when \( BA \) is merged with the frozen pre-trained matrix \( W \), it will result in modifications across all parameters of \( W \). Consequently, this undermines the goal of minimizing changes to the pre-trained model’s knowledge.

We claim that in low-rank adaptation (LoRA), adopting a high sparsity level allows the matrix \( BA \) to remain sparse with high probability. First, we present \cref{theorem:sparsity_expectation}, which provides an expected sparsity estimation for the matrix \( BA \).
\begin{theorem}
\label{theorem:sparsity_expectation}
\vspace{-0.5em}
    Let \( B \in \mathbb{R}^{p \times r} \) and \( A \in \mathbb{R}^{r \times q} \) be two low rank matrices in LoRA, then the expected sparsity of the product matrix \( BA \in \mathbb{R}^{p \times q} \) is given by:
\begin{equation}
\setlength\abovedisplayskip{3pt} \setlength\belowdisplayskip{3pt}
    \mathbb{E}\left[ s_{BA} \right] = 1 - (1 - s_B s_A)^r.
\end{equation}
Proof. See Appendix A.
\hfill\(\Box\)
\vspace{-0.5em}
\end{theorem}

Next, we established an upper bound on the probability that the actual sparsity of  $BA$  exceeds this expectation.
\begin{theorem}
\label{theorem:sparsity_range}
\vspace{-0.5em}
    Let \( B \in \mathbb{R}^{p \times r} \) and \( A \in \mathbb{R}^{r \times q} \) be two low rank matrices in LoRA, where the sparsity of \( B \) is \( s_B \) and the sparsity of \( A \) is \( s_A \). Define \( C = BA \), with sparsity \( s_C \). Then, for any \( \delta > 0 \):
\begin{equation}
\setlength\abovedisplayskip{3pt} \setlength\belowdisplayskip{3pt}
\mathbb{P}\left( \left| s_C - \mathbb{E}[s_C] \right| \geq \delta \right) \leq 2 \exp\left( - \dfrac{2 \delta^2 pq}{r(p + q)} \right),
\end{equation}
where the expected sparsity \( \mathbb{E}[s_C] \) is given by \cref{theorem:sparsity_expectation}
Proof. See Appendix B.
\hfill\(\Box\)
\vspace{-0.5em}
\end{theorem}
It is important to note that these theorems are meaningful only when the rank \( r \) is small, and the sparsity levels \( s_A \) and \( s_B \) are low. Fortunately, in LoRA, matrices \( A \) and \( B \) are inherently low-rank. Additionally, based on prior knowledge of the substantial redundancy in LoRA parameters post-training, the sparsity levels \( s_A \) and \( s_B \) can be set to relatively low values. These conditions ensure the practical of these theorems, suggesting that \( BA \) is sparse with high probability. We further validate these theoretical conclusions with experiments shown in \cref{fig:sparsity_layer}.

\begin{figure*}[t]
\begin{center}
\includegraphics[width=0.9\linewidth]{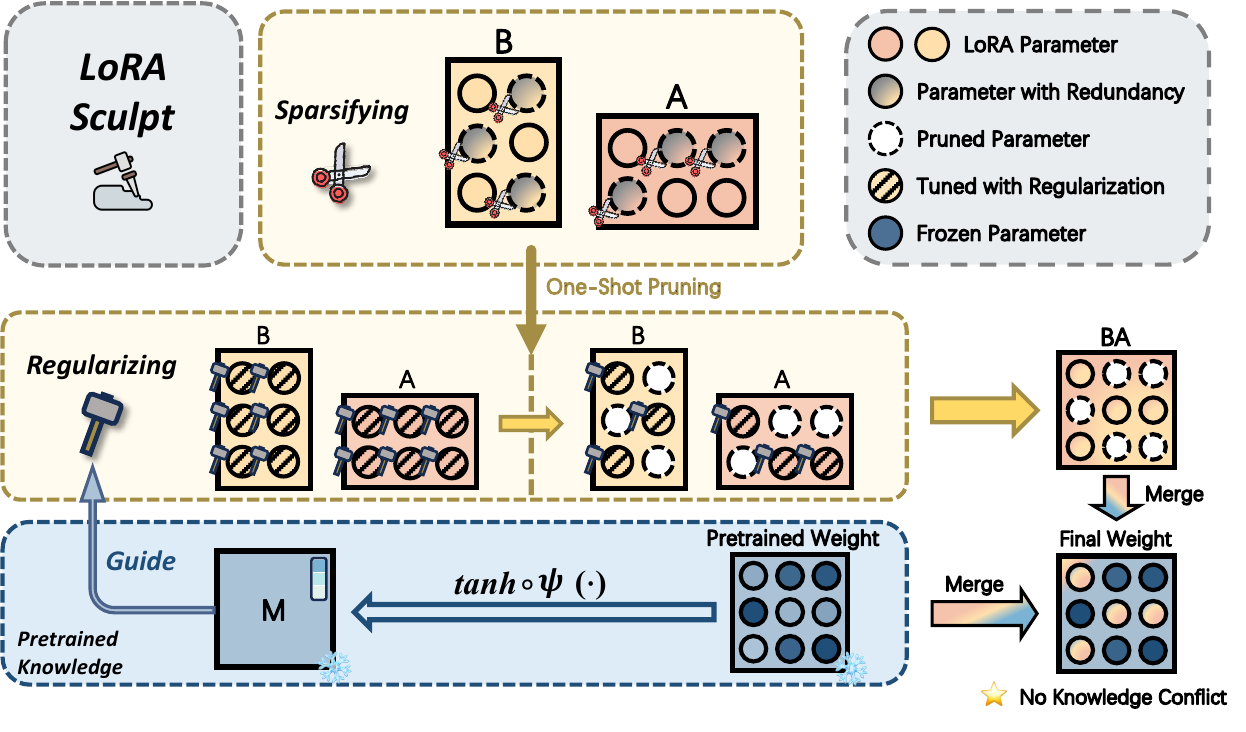}
\put(-278,32){\scriptsize{\cref{eq:cal_s} \cref{eq:cal_m}}}
\put(-335,178){\scriptsize{\cref{eq:sparsify_AB}}} 
\put(-399,105){\scriptsize{\cref{eq:ours_reg}}} 
\end{center}
\vspace{-22pt}
\captionsetup{font=small}
\caption{\small\textbf{Framework Illustration}. LoRASculpt consists of two components: \textit{Sparsifying} and \textit{Regularizing}. \textit{Sparsifying} process aims to reduce redundancy by retaining only a sparse subset of parameters in the low-rank matrices. \textit{Regularizing} process is guided by the pretrained-knowledge informed regularization, which adjusts the optimization trajectory to mitigate conflicts between the sparse LoRA subset and the pretrained knowledge, thereby promoting knowledge harmonization. The resulting sparse product matrix $BA$ can be merged with pretrained weights without severe knowledge conflicts.
}
\vspace{-15pt}
\label{fig:ours}
\end{figure*}

\subsubsection{Regularizing LoRA for Knowledge Harmonization}
\label{sec:Regularizing_LoRA_for_Knowledge_Harmonization}

Under theoretical guarantees, we have successfully introduced sparse updates into LoRA; however, this does not resolve a critical issue: the selected sparse parameter subset may still conflict with important parameters in the pre-trained model. As described in challenge \hyperlink{Q2}{\textbf{\uppercase\expandafter{\romannumeral2})}}, LoRA updates strictly adhere to the empirical loss minimization of the current downstream task, which may leads to interference from pre-trained knowledge behavior during optimization, resulting in the \textbf{\textit{LoRA Optimization Conflict}} and causing general knowledge forgetting.

To mitigate this optimization conflict, we propose \textit{Conflict Mitigation Regularizer} to calibrate the LoRA update trajectory. Specifically, during training, we employ pre-trained knowledge as guidance to regularize LoRA, steering its updates away from the key regions of general knowledge in the pre-trained model and directing new knowledge into less critical areas of the pre-trained parameters. \textit{Notably, it is the high sparsity achievable by LoRA (as mentioned in \cref{sec:Sparsifying_LoRA_for_Redundancy_Reduction}) that enables precise knowledge injection.}

As discussed in previous, the magnitude of pre-trained parameters serves as a crucial measure of importance. Moreover, during efficient fine-tuning with LoRA, gradients of the pre-trained model are inaccessible. Therefore, balancing feasibility and efficiency, we adopt magnitude as the measure of importance of pre-trained weights to generate Magnitude-Guided Retention Mask, as detailed below.

\vspace{0.3em}
\noindent \textbf{Magnitude-Guided Retention Mask.}
To identify the critical components of the pre-trained parameters and guide the subsequent LoRA update trajectory, we apply transformations to the pre-trained parameters to emphasize crucial knowledge, as shown in the following formulation.
\setlength{\abovedisplayskip}{3pt}
\setlength{\belowdisplayskip}{3pt}
\begin{align}
    & S_{ij} = \psi(W_{ij}) = \left| \frac{1}{\log \left( |\widetilde{W}_{ij}| + \epsilon \right)} \right| ,
    \label{eq:cal_s} \\
    & M_{ij} = \tanh(\omega S_{ij}) = \frac{e^{\omega S_{ij}} - e^{-\omega S_{ij}}}{e^{\omega S_{ij}} + e^{-\omega S_{ij}}} ,
    \label{eq:cal_m}
\end{align}
where $\widetilde{W}$ denotes the normalization of pretrained weights, given by $\frac{W}{\|W\|_2}$; \( \epsilon \) denotes a small constant for numerical stability; $\omega$ controls the steepness of the $\tanh$ function, and determines whether weights of different magnitudes are assigned importance values with smooth or sharp distinctions; and \( i = 1, 2, \dots, p \), \( j = 1, 2, \dots, q \) denote matrix indices.

The purpose of the above equations can be divided into two points. \cref{eq:cal_s} alleviates $\log$ function to naturally compress the wide-ranging values of the weight parameters, resulting in a more balanced distribution of importance scores. Parameter with larger magnitude is mapped to higher importance score $S$. \cref{eq:cal_m} rescales the importance scores to the $(0,1)$ range, placing them on a common scale, and facilitating subsequent knowledge-guided regularization. Thus, we obtain the Magnitude-Guided Retention Mask, which represents the strength of protection for the pre-trained knowledge.

\vspace{0.3em}
\noindent \textbf{Conflict Mitigation Regularizer.}
To mitigate the potential conflicts between downstream and essential pre-trained knowledge, which could cause forgetting \cite{mallya2018packnet,kirkpatrick2017overcoming}, we further calibrate the LoRA update trajectory leveraging the retention mask $M$ derived from \cref{eq:cal_m}. 
We propose a pretrained knowledge guided Conflict Mitigation Regularizer by applying the Frobenius norm to the Hadamard product of the retention mask \( M \) and the low-rank LoRA matrix product \( BA \). This reduces the magnitude in \( BA \) corresponding to positions with higher retention strengths in \( M \), leading to smaller modifications to the pre-trained weight \( W \) when mergeing LoRA and thus preserving the essential general knowledge embedded in the pre-trained model. The Conflict Mitigation Regularizer is defined as follows.
\begin{equation}
\setlength\abovedisplayskip{3pt} \setlength\belowdisplayskip{3pt}
    \mathcal{L}_{\textit{\text{CMR}}} = \| M \odot (BA) \|_F
    \text{.}
    \label{eq:reg_cmr}
\end{equation}

Thus, the final loss function is defined as follows.
\begin{equation}
\setlength\abovedisplayskip{3pt} \setlength\belowdisplayskip{3pt}
    \mathcal{L} = \mathcal{L}_{\textit{\text{Task}}} + \alpha \cdot \mathcal{L}_{\textit{\text{CMR}}},
\label{eq:ours_reg}
\end{equation}
where the hyperparameter \( \alpha \) is the balancing coefficient between the two losses, representing the degree of preservation of the pre-trained general knowledge.

\vspace{0.3em}
\noindent\textbf{Synergy Between the Two Components.}
We have proposed two components to address the \textit{LoRA Sparsity Dilemma}, as mentioned in \hyperlink{Q1}{\textbf{\uppercase\expandafter{\romannumeral1})}}, and the \textit{LoRA Optimization Conflict}, as discussed in \hyperlink{Q2}{\textbf{\uppercase\expandafter{\romannumeral2})}}. These components work synergistically toward a common goal: eliminating redundancy and identifying a critical sparse subset to alleviate knowledge conflicts.
Specifically, we implement one-shot pruning during training, maintaining this sparse subset unchanged \footnote{Crucial for fine-tuning with an unchanged sparse subset \cite{DiffPurning_ACL20}}, and divide the training process into two phase: \textit{knowledge warming} phase and \textit{knowledge integration} phase. The proposed knowledge-guided regularization effectively functions in both phases: in the knowledge warming phase, it steers LoRA away from conflicts with critical pre-trained knowledge, aiding in the selection of the sparse parameter subset. In the knowledge integration phase, it leverages pre-trained knowledge to guide the optimization process, further harmonizing the selected sparse parameter subset with pre-trained knowledge to promote knowledge integration.

\vspace{0.3em}
\noindent \textbf{Theoretical Proof of Sparsity.}
In \cref{theorem:sparsity_expectation} and \cref{theorem:sparsity_range}, we have demonstrated the expected sparsity of the resulting delta weight after introducing sparse updates into LoRA and further establish an upper bound for the probability that the actual sparsity exceeds this expectation. Furthermore, we claim that with the application of Knowledge-Guided Regularization, both the expected sparsity and its bound remain valid. Please refer to Appendix D for detailed proofs.

\subsubsection{Method Adaptation in MLLM Connector}
The connector is a key bridge between the visual encoder and language model in multimodal large language models, with the quality of conveyed visual information directly impacting overall performance \cite{Honeybee_CVPR24,li2024tokenpacker}. 
Therefore, full fine-tuning is often employed to connector when adapting MLLMs on downstream tasks \cite{LLaVA_NeurIPS23,LLaVA15_CVPR24,PEFTMLLM_ACL24}.

Our experiment shown in \cref{tab:adapt_connector} reveals that, despite the connector having much fewer parameters than the LLM, it still experiences substantial forgetting. Given the connector’s key role in alignment, we make a simple modification to \ourmethodslim{} for its adaptation in connector. Specifically, to prevent potential performance degradation, we avoid hard pruning on the connector and instead apply regularization for soft sparsity, as shown in the following equation.
\begin{equation}
\setlength\abovedisplayskip{3pt} \setlength\belowdisplayskip{3pt}
    \mathcal{L}_{\textit{\text{CMR}}}^{\text{Con}} = \| M^{\text{Con}} \odot (BA) \|_1,
    \label{eq:loss_conn}
\end{equation}
where \(\| \cdot \|_1\) is \(L_1\) norm, which encourages soft sparsity in $BA$.
Thus, the final loss function when applying \ourmethodslim{} into 
MLLM connector is defined as follows.
\begin{equation}
\setlength\abovedisplayskip{3pt} \setlength\belowdisplayskip{3pt}
    \mathcal{L} = \mathcal{L}_{\textit{\text{Task}}} + \alpha \cdot \mathcal{L}_{\textit{\text{CMR}}}^{\text{LLM}} + \beta \cdot \mathcal{L}_{\textit{\text{CMR}}}^{\text{Con}},
\label{eq:adapt_connector}
\end{equation}
where $\alpha$ and $\beta$ represent the degree of preservation of pre-trained general knowledge in the LLM and the connector, respectively; $\mathcal{L}_{\textit{\text{CMR}}}^{\text{LLM}}$  is consistent with the formula in \cref{eq:reg_cmr}.

The framework of \ourmethod{} is illustrate in \cref{fig:ours}, and the algorithm is outlined in Appendix E.

\section{Experiments}

\subsection{Experimental Setup}

\begin{table*}[t]\small
\centering
\scriptsize{
\resizebox{\linewidth}{!}{
\setlength\tabcolsep{3.pt}
\renewcommand\arraystretch{1.1}
\begin{tabular}{r||cccc|cc|cIcccc|cc|c}
\hline\thickhline
\rowcolor{gray!20}
 & 
\multicolumn{7}{cI}{\bm{\iconqa{}}} & 
\multicolumn{7}{c}{\bm{{\coco{}}}}
\\
\cline{2-15} 
\rowcolor{gray!20}
\multirow{-2}{*}{Methods\quad}  
& \okvqa{} & \ocrvqa{} & \gqa{} & \textvqa{} & \textit{Source} & \textit{Target}  & \textit{Avg}
& \okvqa{} & \ocrvqa{} & \gqa{} & \textvqa{} & \textit{Source} & \textit{Target}  & \textit{Avg}
\\
\hline\hline

\zeroshot{} 
& 57.99 & 66.20 & 61.93 & 58.23 & 61.09 & 23.18  & 42.13 
& 57.99 & 66.20 & 61.93 & 58.23 & 61.09 & 40.40  & 50.74 
\\
\hline\hline

\multicolumn{14}{l}{\textcolor{gray!80}{\textit{LoRA Rank=16}}}

\\
\rowcolor{gray!10} \lora{} 
& 51.10 & 54.90 & 55.56 & 49.48 & 52.76 & 84.14  & 68.45   
& 47.34 & 60.55 & 56.91 & 45.61 & 52.60 & 112.22  & 82.41 
\\

\dora{} 
& 49.94 & 54.40 & 55.39 & 47.26 & 51.75 & 84.48 & 68.11 
& 45.38 & 59.75 & 55.40 & 41.38 & 50.48 & 112.60 & 81.54 
\\

\rowcolor{gray!10} \ltworeg{} 
& 49.96 & 51.40 & 55.15 & 47.47 & 51.00 & \underline{84.66} & 67.83   
& 46.05 & 61.50 & 56.60 & 43.54 & 51.92 & 111.83 & 81.88 
\\

\orthreg{} 
& 53.12 & 56.10 & 57.43 & 51.00 & \underline{54.41} & 84.52 & \underline{69.47} 
& 48.11 & 59.85 & 56.83 & 46.69 & 52.87 & 111.87 & 82.37 
\\

\rowcolor{gray!10} \tailor{} 
& 52.68 & 56.55 & 56.26 & 51.41 & 54.23 & 83.26 & 68.74  
& 51.83 & 59.75 & 58.73 & 51.43 & \textbf{55.44} & \underline{118.84} & \underline{87.14} 
\\

\dare{}
& 51.10 & 54.30 & 55.55 & 49.20 & 52.54 & 84.20 & 68.37  
& 46.64 & 59.45 & 56.01 & 44.46 & 51.64 & 112.21 & 81.93  
\\
\hline
\rowcolor[HTML]{D7F6FF}
\ourmethod{}{}
& 54.07 & 57.00 & 57.66 & 52.24 & \textbf{55.24} & \textbf{85.02} & \quad\textbf{70.13}\redup{0.66}\quad  
& 50.86 & 59.10 & 57.64 & 51.10 & \underline{54.68} & \textbf{121.26} & \quad\textbf{87.97}\redup{0.83}\quad  
\\
\hline\hline

\multicolumn{14}{l}{\textcolor{gray!80}{\textit{LoRA Rank=32}}}

\\
\rowcolor{gray!10} \lora{} 
& 45.06 & 48.40 & 49.22 & 36.47 & 44.79 & 82.52 & 63.65   
& 44.92 & 59.90 & 54.88 & 39.75 & 49.86 & 110.27 & 80.07 
\\

\dora{} 
& 46.35 & 48.50 & 40.80 & 33.01 & 42.17 & \underline{84.82} & 63.49 
& 44.10 & 60.05 & 54.69 & 40.17 & 49.75 & 109.25 & 79.50 
\\

\rowcolor{gray!10} \ltworeg{} 
& 30.16 & 26.65 & 16.62 & 17.16 & 22.65 & 82.36 & 52.50   
& 43.60 & 60.00 & 54.09 & 36.53 & 48.56 & 109.39 & 78.97 
\\

\orthreg{} 
& 41.51 & 40.45 & 53.58 & 30.92 & 41.62 & 83.14 & 62.38 
& 43.77 & 60.70 & 53.74 & 38.61 & 49.21 & 110.44 & 79.82 
\\

\rowcolor{gray!10} \tailor{} 
& 50.07 & 53.85 & 53.04 & 46.92 & \underline{50.97} & 81.76 & \underline{66.37}  
& 50.99 & 60.60 & 58.54 & 49.65 & \textbf{54.95} & \underline{117.64} & \underline{86.29} 
\\

\dare{}
& 44.39 & 46.85 & 48.75 & 34.84 & 43.71 & 82.28 & 62.99  
& 42.98 & 58.65 & 53.92 & 38.61 & 48.54 & 108.57 & 78.56  
\\
\hline
\rowcolor[HTML]{D7F6FF}
\ourmethod{}{}
& 53.52 & 59.50 & 57.63 & 53.76 & \textbf{56.10} & \textbf{85.26} & \quad\textbf{70.68}\redup{4.31}\quad   
& 49.99 & 58.65 & 57.63 & 50.73 & \underline{54.25} & \textbf{120.35} & \quad\textbf{87.30}\redup{1.01}\quad  
\\
\hline\hline

\multicolumn{14}{l}{\textcolor{gray!80}{\textit{LoRA Rank=64}}}

\\
\rowcolor{gray!10} \lora{} 
& 0.04 & 0.00 & 20.22 & 0.28 & 5.14 & 37.24 & 21.19   
& 31.51 & 54.80 & 42.35 & 28.13 & 39.20 & 107.90 & 73.55 
\\

\dora{} 
& 0.03 & 0.00 & 0.00 & 0.12 & 0.04 & 36.86 & 18.45 
& 34.63 & 57.85 & 44.37 & 29.06 & 41.48 & 106.49 & 73.98 
\\

\rowcolor{gray!10} \ltworeg{} 
& 27.46 & 31.75 & 17.98 & 9.66 & 21.71 & \underline{81.79} & 51.75   
& 40.95 & 59.05 & 52.83 & 36.00 & 47.21 & 106.41 & 76.81 
\\

\orthreg{} 
& 39.04 & 35.90 & 25.59 & 21.21 & 30.44 & 81.71 & 56.07 
& 45.78 & 60.70 & 55.69 & 43.58 & 51.44 & 108.16 & 79.80 
\\

\rowcolor{gray!10} \tailor{} 
& 47.67 & 51.50 & 52.61 & 35.43 & \underline{46.80} & 77.28 & \underline{62.04}  
& 46.70 & 59.20 & 56.75 & 44.42 & \underline{51.77} & \underline{116.68} & \underline{84.22} 
\\

\dare{}
& 34.42 & 39.65 & 21.90 & 14.59 & 27.64 & 78.58 & 53.11  
& 29.67 & 50.70 & 37.54 & 23.34 & 35.31 & 105.55 & 70.43  
\\
\hline
\rowcolor[HTML]{D7F6FF}
\ourmethod{}{}
& 53.71 & 57.90 & 57.60 & 52.14 & \textbf{55.34} & \textbf{84.74} & \quad\textbf{70.04}\redup{8.00}\quad   
& 50.15 & 59.85 & 57.91 & 51.44 & \textbf{54.84} & \textbf{120.39} & \quad\textbf{87.61}\redup{3.45}\quad
\\
\hline
\end{tabular}}}
\vspace{-5pt}
\captionsetup{font=small}
\caption{{
\textbf{Comparison with State-of-the-Art Fine-Tuning Solutions for Multimodal Large Language Models (MLLMs)} on visual question answering task \iconqa{} and image captioning task \coco{}.
The optimal and sub-optimal results are denoted by boldface and underlining. {\color{RedOrange}$\uparrow$} means improved accuracy compared with the sub-optimal results. Please refer to \cref{sec:comparison_to_sota} for detailed explanations.
}}
\label{tab:compare_sota_image_capation}
\vspace{-10pt}
\end{table*}

\begin{table}[t]\small
\centering
{
\resizebox{\columnwidth}{!}{
\setlength\tabcolsep{5pt}
\renewcommand\arraystretch{1.1}
\begin{tabular}{cc||cccc|ccc}
\thickhline
\rowcolor{lightgray} &
& \multicolumn{7}{c}{\textbf{Fine-Tune on} \bm{\iconqa{}}} \\
\cline{3-9} 
\rowcolor{lightgray} 
\multirow{-2}{*}{LLM} & \multirow{-2}{*}{Conn}  
& \okvqa{} & \ocrvqa{} & \gqa{} & \textvqa{} & \textit{Source} & \textit{Target}  & \textit{Avg} \\
\hline\hline
\multicolumn{2}{c||}{\zeroshot{}} 
& 57.99 & 66.20 & 61.93 & 58.23
& 61.09 & 23.18 & 42.13
\\ 
\hdashline
\lora{} & FFT
& 0.04 & 0.00 & 20.22 & 0.28
& 5.14 & 37.24 & 21.19 
\\ 
\lora{} & \lora{}
& 0.76 & 2.45 & 0.01 & 2.48
& 1.43 & 81.60 & 41.51
\\ 
\lora{} & Freeze
& 51.10 & 52.15 & 54.17 & 45.91
& 50.83 & 84.61 & 67.72
\\
\rowcolor{gray!10}
\textbf{Ours} & \textbf{FFT}
& 53.71 & 57.90 & 57.60 & 52.14
& 55.34 & 84.74 & 70.04
\\
\rowcolor[HTML]{D7F6FF}
\textbf{Ours} & \textbf{Ours}
& 56.91 & 64.15 & 60.93 & 56.22
& \textbf{59.55} & \textbf{85.34} & \textbf{72.45}
\\
\hline
\end{tabular}}}
\vspace{-5pt}
\captionsetup{font=small}
\caption{
\textbf{Method Adaptation in MLLM Connector} shows significant catastrophic forgetting within the MLLM connector. Please refer to \cref{sec:cf_connector} for more details.
}
\label{tab:adapt_connector}
\vspace{-5pt}
\end{table}

\noindent\textbf{Datasets and Architecture.}
For downstream knowledge acquisition, we fine-tune and evaluate on Visual Question Answering (VQA) and Captioning tasks using the \iconqa{} \cite{IconQA_NeurIPS21} and \coco{} \cite{COCO_ECCV14} datasets. We randomly sample 10k examples from each training dataset, following the resource setting in {\cite{ReviEFT_arXiv22,PEFTMLLM_ACL24}}. To measure general knowledge forgetting in MLLM, we assess performance on four upstream datasets: OKVQA \cite{OKVQA_CVPR19}, OCRVQA \cite{OCRVQA_ICDAR19}, GQA \cite{GQA_CVPR19}, and TextVQA \cite{TextVQA_CVPR19} after downstream task fine-tuning. Leveraging the representative open-source framework LLaVA-1.5 \cite{LLaVA15_CVPR24}, we conduct detailed evaluations of compared LoRA-based methods at ranks 16, 32, and 64. 

\noindent\textbf{Implementation Details.} 
All compared baselines are adapted based on LoRA and implemented within the LLaVA-1.5\footnote{https://github.com/haotian-liu/LLaVA} framework. Consistent with several works \cite{LLaVA15_CVPR24,PEFTMLLM_ACL24}, we apply LoRA to all layers of LLM with a learning rate of 2e-4, and use full fine-tuning in the connector with a learning rate of 2e-5.
The training epoch is set to 3, and the batch size is default set to 16. 
Regularizing LoRA for Knowledge Harmonization (RKH) is only applied to $W_Q, W_K, W_V$, to encourage higher sparsity and knowledge harmonization, as attention layers have been shown to be more significant and redundant than others \cite{he2024matters}.
All experiments are conducted on 4 NVIDIA 4090 GPUs. 

\noindent\textbf{Compared Baselines.}
We compare proposed method \ourmethodslim{} against LoRA-based PEFT methods, regularization techniques, and post-training approaches, including: 
(a) {\lora{}} \pub{ICLR'22} \cite{LoRA_ICLR22} 
(b) {\dora{}} \pub{ICML'24} \cite{DoRA_ICML24} 
(c) {\orthreg{}} \pub{ECCV'24} \cite{PEGO_ECCV24} 
(d) {L2-Regularization} \pub{PNAS'17} \cite{EWC_PNAS17} 
(e) {\dare{}} \pub{ICML'24} \cite{DARE_ICML24} 
(f) {\tailor{}} \pub{ICML'24} \cite{ModelTailor_ICML24}. 
Additional evaluation details are provided in Appendix F.

\begin{table}[t]\small
\centering
{
\resizebox{\columnwidth}{!}{
\setlength\tabcolsep{5pt}
\renewcommand\arraystretch{1.1}
\begin{tabular}{cc||cccIccc}
\hline \thickhline
\rowcolor{lightgray} &
& \multicolumn{3}{cI}{\bm{\iconqa{}}} 
& \multicolumn{3}{c}{\bm{\coco{}}} \\
\cline{3-8} 
\rowcolor{lightgray}
\multirow{-2}{*}{SRR} & \multirow{-2}{*}{RKH}  
&  \textit{Source} & \textit{Target} & \textit{Avg}
&  \textit{Source} & \textit{Target} & \textit{Avg}
\\
\hline\hline
\multicolumn{2}{c||}{\zeroshot{}} 
& 61.09 & 23.18 & 42.13
& 61.09 & 40.40 & 50.74
\\ 
\hdashline
\multicolumn{2}{c||}{\lora{}} 
& 44.79 & 82.52 & 63.65  
& 49.86 & 110.27 & 80.07 
\\ 
\ding{51} & 
& 54.73 & 84.45 & 69.59    
& 53.97 & 120.11 & 87.04
\\
\rowcolor[HTML]{D7F6FF}
\ding{51} & \ding{51}
& \textbf{56.10} & \textbf{85.26} & \textbf{70.68}  
& \textbf{54.25} & \textbf{120.35} & \textbf{87.30}  
\\
\hline
\end{tabular}}}
\vspace{-5pt}
\captionsetup{font=small}
\caption{
\textbf{Ablative Study of Key Modules} for LoRASculpt, showing the effects of Sparsifying for Redundancy Reduction (SRR) and Regularizing for Knowledge Harmonization (RKH). Please refer to \cref{sec:ablation_study} for detailed discussion.
}
\label{tab:ablation_module}
\vspace{-15pt}
\end{table}

\subsection{Comparison to State-of-the-Arts}
\label{sec:comparison_to_sota}
\textbf{Quantitative Results.}
We conduct extensive experiments comparing \ourmethodslim{} with state-of-the-art baselines across different ranks on two downstream tasks, as shown in \cref{tab:compare_sota_image_capation}. Several key observations are summarized:

\ding{182} \textbf{\textit{LoRASculpt alleviates the trade-off between general and specialized knowledge.}} By sculpting LoRA with redundancy reduction and knowledge harmonization, it achieves better results on both \textit{target} and \textit{source} compared to baseline LoRA methods, consistently achieving state-of-the-art results across different ranks (16, 32, 64) and tasks.

\ding{183} \textbf{\textit{Existing methods carry the risk of reducing downstream task accuracy.}} Excluding cases of catastrophic forgetting at rank=64 on IconQA (discussed in \cref{sec:cf_connector}), both regularization and post-pruning methods may underperform compared to the LoRA baseline. For instance, regularization methods generally show lower target performance on the \coco{} dataset, and post-pruning methods show similar results on \iconqa{}.

\ding{184} \textbf{\textit{LoRASculpt demonstrates great robustness.}} As the LoRA rank increases, the risk of forgetting intensifies, and existing methods show similar declines in performance. In contrast, LoRASculpt consistently maintains stable results.

\begin{figure}[t]
\centering
\includegraphics[width=0.98\linewidth]{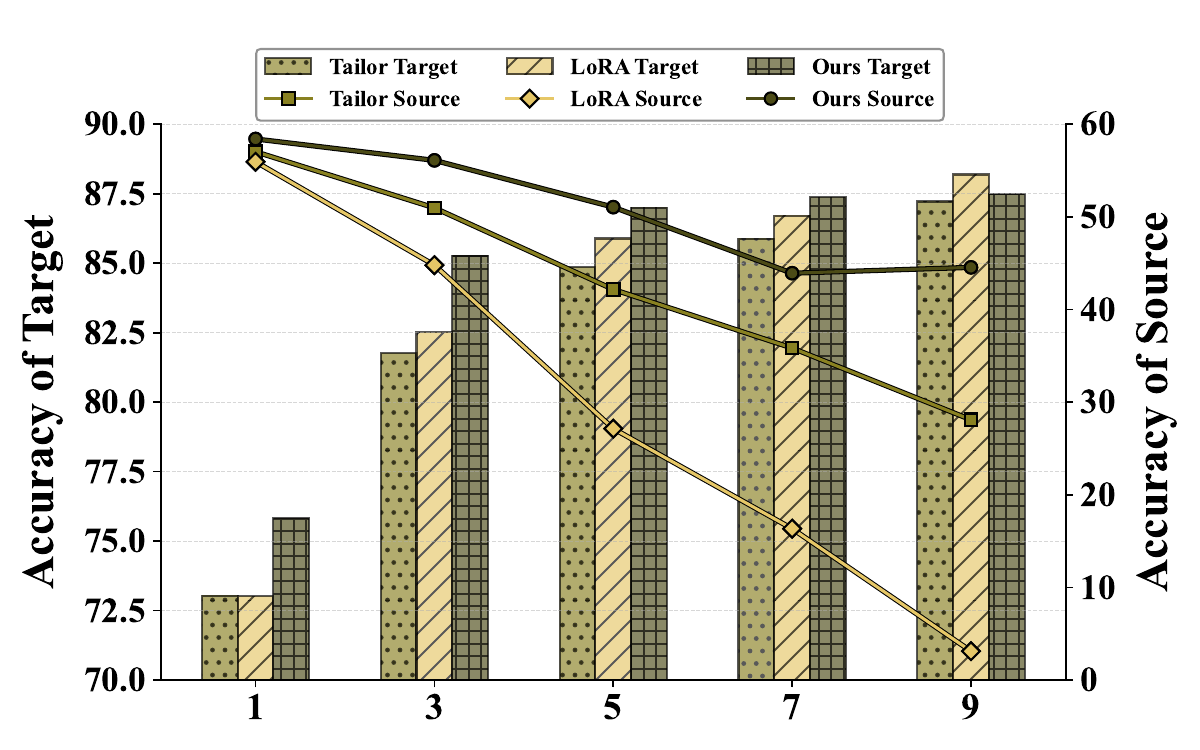}
\vspace{-10pt}
\captionsetup{font=small}
\caption{\textbf{Results Across Different Epochs} compared with SOTA method and baseline, showing LoRA obvious decline in \textit{source} and Tailor compromise in \textit{target} performance, while our method maintains stable for both. Please see details in \cref{sec:comparison_to_sota}
}
\label{fig:epoch_exp}
\vspace{-15pt}
\end{figure}

\noindent\textbf{Results Across Different Epochs.} As SFT progresses, forgetting issue tends to worsen over longer training durations \cite{biderman2024lora}. Thus, we examine the results of different methods as training epochs increase on \iconqa{} dataset. 
\cref{fig:epoch_exp} shows how \textit{target} and \textit{source} accuracy change across epochs for Tailor \cite{ModelTailor_ICML24}, LoRA \cite{LoRA_ICLR22}, and our method. As epochs increase, LoRA display a steady decline in \textit{source} accuracy, reflecting worsening forgetting. Although Tailor mitigates forgetting, it compromises downstream performance. In contrast, our method maintains stable accuracy for both \textit{target} and \textit{source}, demonstrating greater robustness.

\subsection{Diagnostic Analysis}
\label{sec:Diagnostic_Analysis}

\subsubsection{Catastrophic Forgetting in MLLM Connector}
\label{sec:cf_connector}

Increasing the rank of LoRA may raise the risk of overfitting and catastrophic forgetting \cite{biderman2024lora,wang2024lora}. In our experiments in \cref{tab:compare_sota_image_capation}, we observed that when LoRA rank is set to 64 and fine-tuned for 3 epochs, both LoRA and DoRA without forgetting mitigation technique exhibit catastrophic forgetting. Our further experiments in \cref{tab:adapt_connector} indicate that this issue is \textit{\textbf{largely due to catastrophic forgetting within the MLLM connector.}} The results show that applying LoRA to the connector alleviates overfitting but still leads to catastrophic forgetting, while freezing the connector significantly reduces this effect. Additionally, our experiments demonstrate that applying our method to the connector with \cref{eq:adapt_connector} achieves a better harmonization between general and specialized knowledge.
For ablation study on the parameter $\beta$, which controls the sparsity strength for MLLM connector, please refer to the Appendix G.

\subsubsection{Ablation Study}
\label{sec:ablation_study}

\noindent\textbf{Ablation of Key Component.}
We present an ablation study of two key components of LoRASculpt in \cref{tab:ablation_module}: including Sparsifying for Redundancy Reduction (SRR) and Regularizing for Knowledge Harmonization (RKH). Our experiments demonstrate the effectiveness of SRR in removing harmful redundant parameters in LoRA, while RKH further enhances the harmonization between the general and downstream task knowledge.

\begin{figure}[t]
    \centering 

    \begin{subfigure}{0.49\columnwidth}
        \centering
        \includegraphics[width=\linewidth]{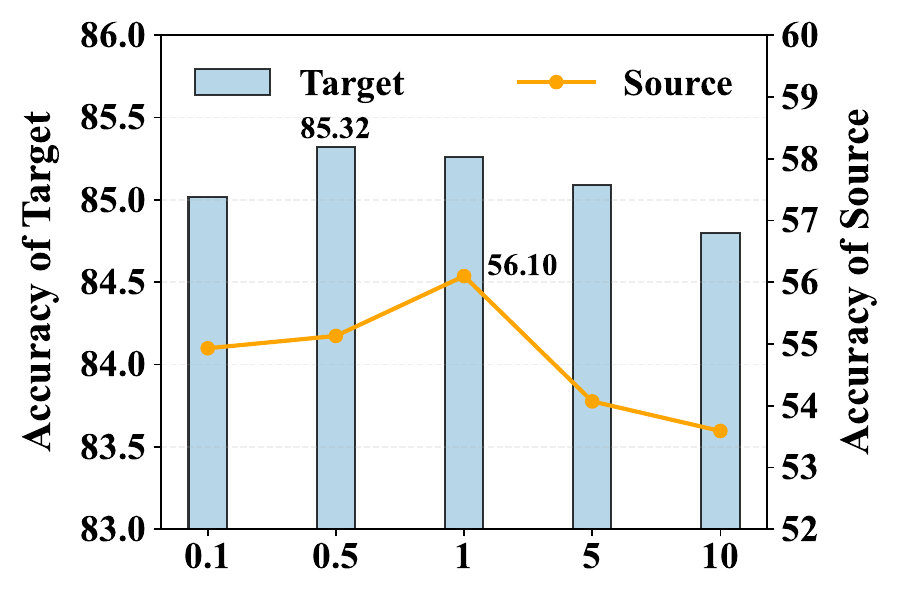} 
        \caption{Ablation of $\omega$}
        \label{fig:omega_ablation}
    \end{subfigure}
    \hfill
    \begin{subfigure}{0.49\columnwidth}
        \centering
        \includegraphics[width=\linewidth]{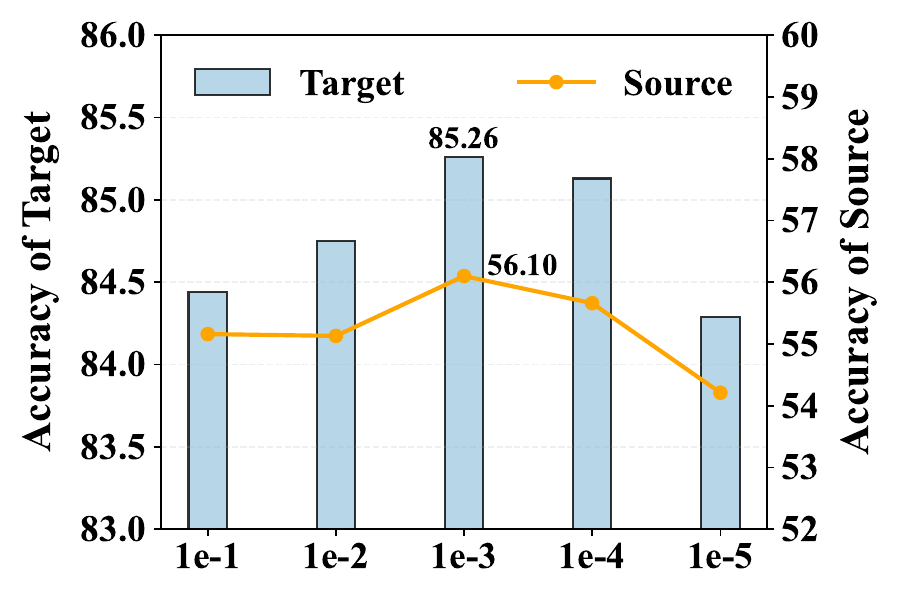} 
        \caption{Ablation of $\alpha$}
        \label{fig:alpha_ablation}
    \end{subfigure}
    
    \vspace{0.03cm} 
    \begin{subfigure}{0.95\columnwidth} 
        \centering
        \includegraphics[width=\linewidth]{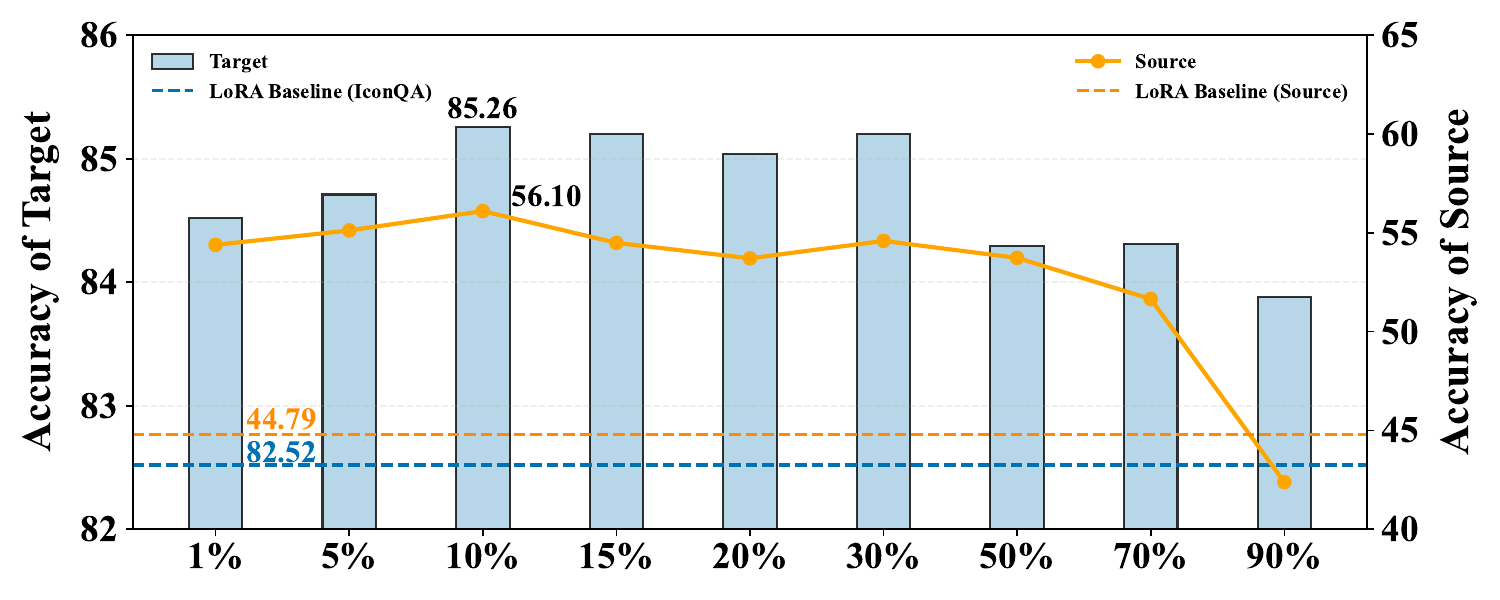}
        \caption{Ablation of Sparsity Ratio $s$}
        \label{fig:ratio_ablation_with_baseline}
    \end{subfigure}
    \vspace{-5pt}
    \caption{\textbf{Hyperparameter Study} for function steepness $w$ (\cref{eq:cal_m}), balancing coefficient $\alpha$ (\cref{eq:ours_reg}), and sparsity ratio $s$ (\cref{eq:sparsify}) when fix LoRA rank=32 and fine-tuning on \iconqa{}. Please refer to \cref{sec:ablation_study} for detailed discussion.}
    \label{fig:hyper_ablations}
    \vspace{-11pt}
\end{figure}

\noindent\textbf{Ablation of Hyperparameters.}
We show the impact of the hyperparameters $\alpha$ (\cref{eq:ours_reg}) and $\omega$ (\cref{eq:cal_m}) on the performance in \cref{fig:alpha_ablation} and \cref{fig:omega_ablation}. We set the sparsity ratio of low-rank matrices $A$ and $B$ as $s_A=s_B=s$ (\cref{eq:sparsify}), then evaluate performance across different sparsity in \cref{fig:ratio_ablation_with_baseline}. Optimal performance is achieved when $\omega=1$, $\alpha=10^{-3}$ and $s_A=s_B=s=10\%$,  which are used in main experiments \cref{tab:compare_sota_image_capation}. Further analysis of the figure reveals that when the sparsity ratio is set to 1\%, performance on both source and target surpasses the LoRA baseline, indicating that LoRA contains highly redundancy after fine-tuning on MLLMs. This aligns with our observations in \cref{sec:Preliminary}.

\subsubsection{Empirical Validation of Proposed Theorem}
\label{sec:Empirical_Validation}

\begin{figure}[t]
    \centering 

    \begin{subfigure}{0.48\columnwidth}
    \includegraphics[width=\linewidth]{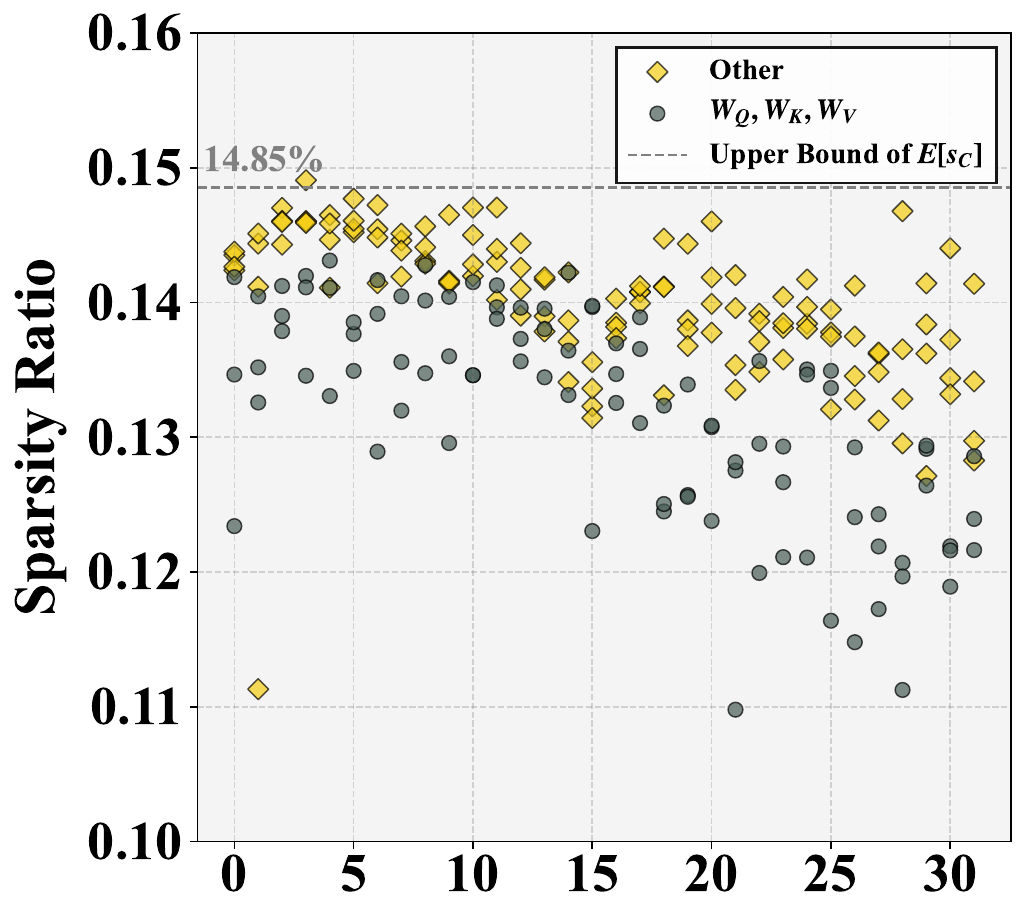}
    \caption{LoRA \textit{r=16} on IconQA}
    \end{subfigure}
    \hspace{-5pt}
    \begin{subfigure}{0.447\columnwidth}
    \includegraphics[width=\linewidth]{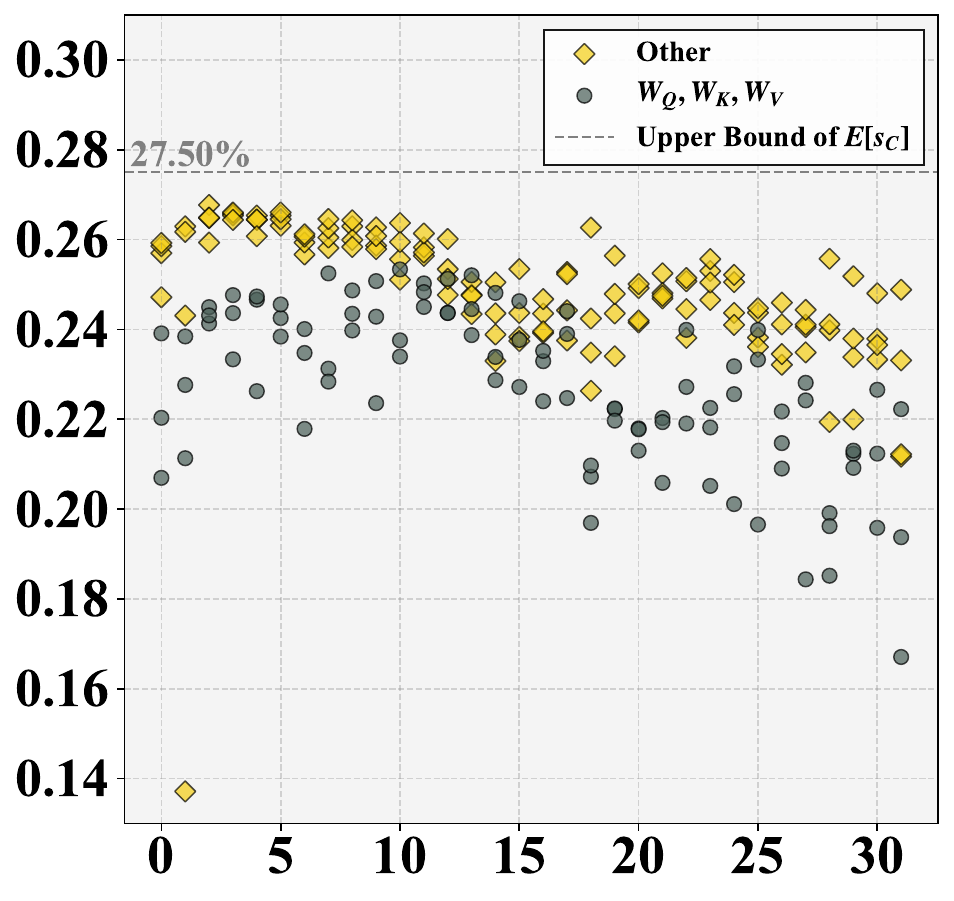}
    \caption{LoRA \textit{r=32} on IconQA}
    \end{subfigure}
    
    \begin{subfigure}{0.48\columnwidth}
    \includegraphics[width=\linewidth]{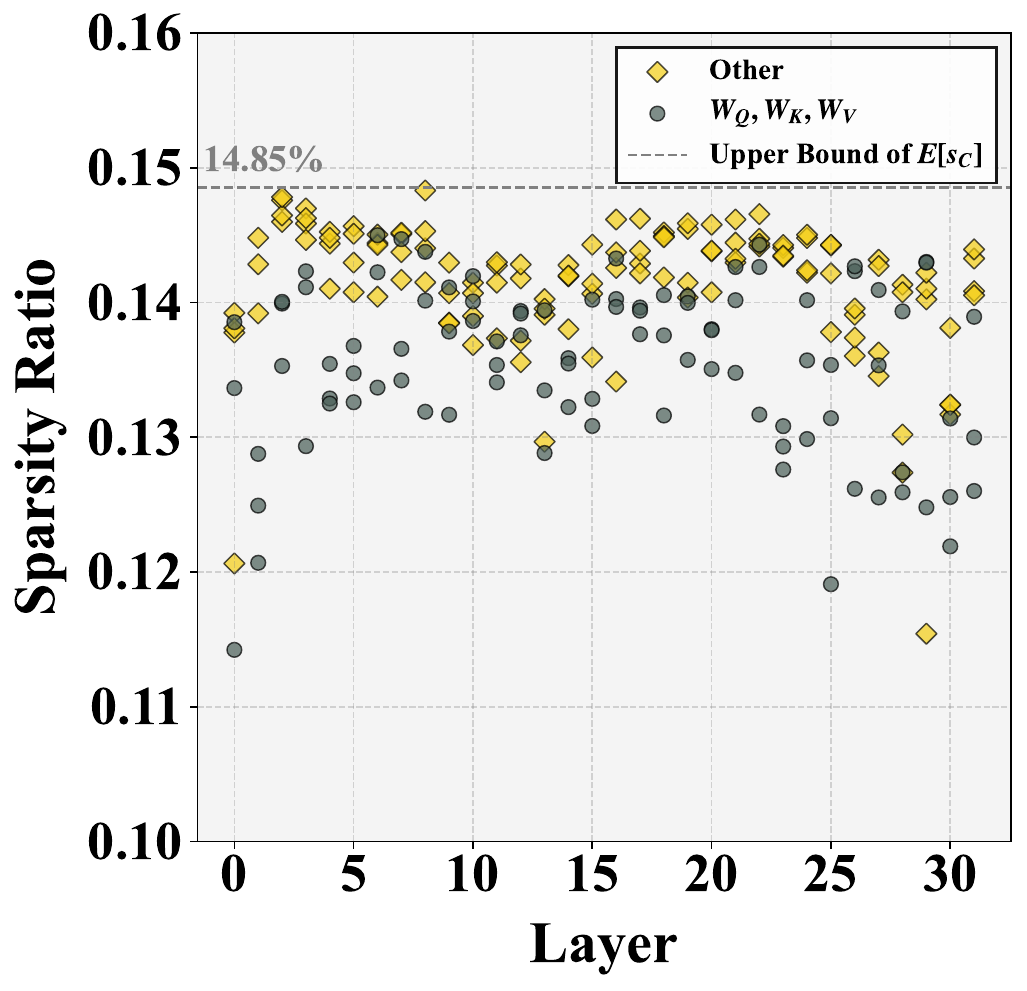}
    \caption{LoRA \textit{r=16} on COCO}
    \end{subfigure}
    \hspace{-5pt}
    \begin{subfigure}{0.447\columnwidth}
    \includegraphics[width=\linewidth]{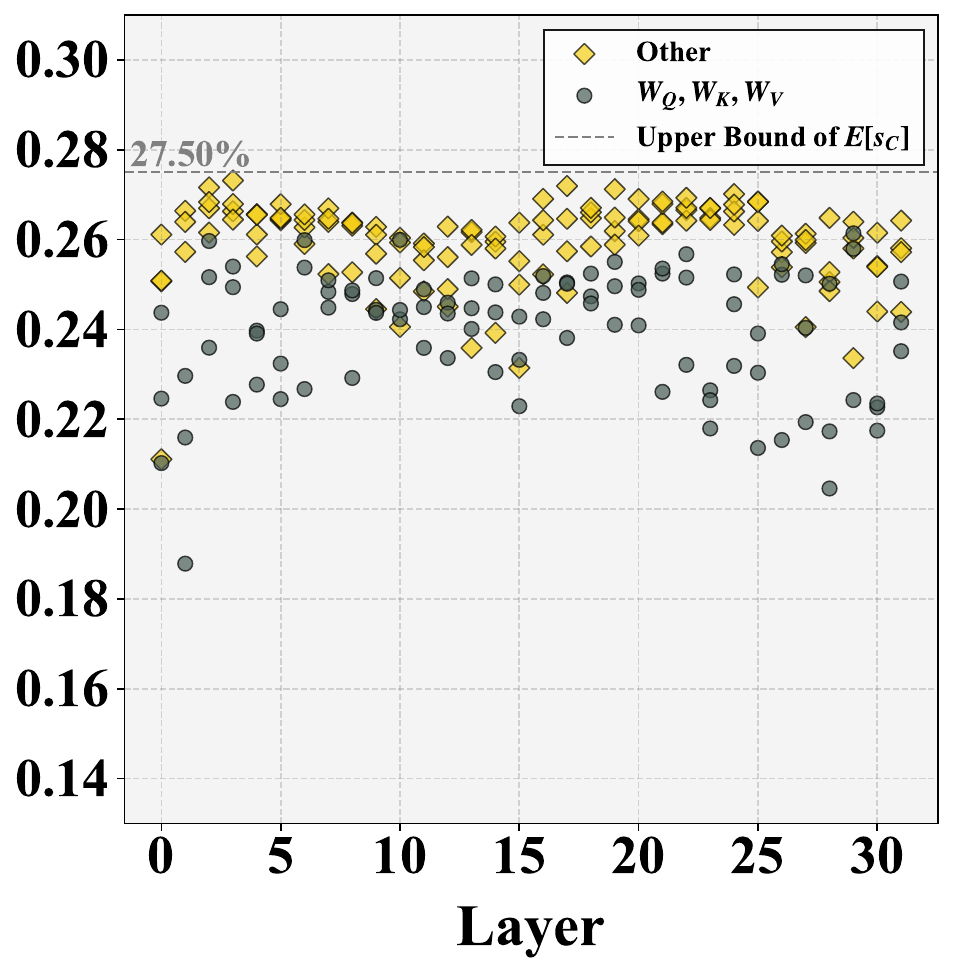}
    \caption{LoRA \textit{r=32} on COCO}
    \end{subfigure}
    
    \vspace{-5pt}
    \caption{\small{
    \textbf{Actual Sparsity of LoRA in Each Layer} with the sparsity $s_A=s_B=0.1$. Please refer to \cref{sec:Empirical_Validation} for details.
    }}
    \label{fig:sparsity_layer}
    \vspace{-12pt}
\end{figure}

In \cref{sec:Proposed_Method}, we prove an upper bound (\cref{theorem:sparsity_expectation}) on the expected sparsity of $BA$ given fixed sparsity levels in the low-rank matrices $A$ and $B$, and further provide an upper bound for the probability that actual sparsity exceeds this expectation (\cref{theorem:sparsity_range}). We present experimental validation shown in \cref{fig:sparsity_layer},  where the sparsity of the product matrix $BA$ in nearly all layers falls below the expected sparsity derived from theoretical analysis (indicated by the gray line). Under these theoretical guarantees, our method successfully introduces sparsity into LoRA.

Further analysis of \cref{fig:sparsity_layer} reveals two insights: (1) The sparsity of LoRA in the MLP layers ($11008 \times 4096$) is closer to the expected bound than in the $W_Q, W_K, W_V$ layers ($4096 \times 4096$), aligning with the conclusion of \cref{theorem:sparsity_range}. (2) LoRASculpt applies sparsity to the low-rank matrices B and A, automatically enforcing varying degrees of redundancy reduction across layers. Recent studies on MLLM layer-wise roles suggest that shallow layers primarily integrate global image information, while deeper layers handle syntactic structures \cite{zhang2024cross,huang2025keeping,zhu2025remedy}. For IconQA dataset, where answers are simple single letters, LoRASculpt exhibits lower sparsity in deeper layers. Conversely, for the COCO-Caption dataset, which requires diverse answer formats, deeper layers retain more parameters to capture rich semantic information. This demonstrates the adaptive redundancy reduction of LoRASculpt rather than applying a fixed sparsity rate across all layers.

\section{Conclusion}
In this paper, we introduce \ourmethod{}, a parameter-efficient framework designed to address the forgetting issue for MLLMs. 
Our method consists of two components: \textit{Sparsifying LoRA for Redundancy Reduction}, which eliminates harmful redundancy, and \textit{Regularizing LoRA for Knowledge Harmonization}, which balances general and task-specific knowledge to achieve better alignment. 
Through theoretical analysis and extensive experiments, we show that our method effectively retains general knowledge while enhancing downstream task performance.

\vspace{0.5em}
\noindent \textbf{Acknowledgement.}
This research is supported by the National Key Research and Development Project of China (2024YFC3308400), the National Natural Science Foundation of China (Grants 62361166629, 62176188, 623B2080), the Wuhan University Undergraduate Innovation Research Fund Project. The supercomputing system at the Supercomputing Center of Wuhan University supported the numerical calculations in this paper.

\clearpage
\setcounter{page}{1}
\maketitlesupplementary
\appendix
\renewcommand{\thetable}{\Roman{table}}
\setcounter{table}{0}

\section{Proof of Theorem 3.1}

\noindent\textbf{Theorem 3.1.} 
\textit{Let \( B \in \mathbb{R}^{p \times r} \) and \( A \in \mathbb{R}^{r \times q} \) be two low rank matrices in LoRA, then the expected sparsity of the product matrix \( BA \in \mathbb{R}^{p \times q} \) is given by:}
\begin{equation}
\setlength\abovedisplayskip{3pt} \setlength\belowdisplayskip{3pt}
    \mathbb{E}\left[ s_{BA} \right] = 1 - (1 - s_B s_A)^r.
\end{equation}

\vspace{-0.8em}
\begin{proof}
We aim to determine the expected proportion of non-zero elements in the product matrix \( BA \in \mathbb{R}^{p \times q} \). The element in the \( i \)-th row and \( j \)-th column of \( BA \) is given by
\begin{equation}
    (BA)_{ij} = \sum_{k=1}^{r} B_{ik} A_{kj}.
\end{equation}

We will prove a stronger conclusion: we assume that all elements in \( A \) and \( B \) are nonnegative. This assumption increases the number of nonzero elements in  $BA$, making it more challenging to ensure the sparsity of  $BA$.

Then an element \( (BA)_{ij} \) is non-zero if and only if there exists at least one \( k \in \{1, 2, \dots, r\} \) such that both \( B_{ik} \) and \( A_{kj} \) are non-zero.
For each \( k \), the probability that \( B_{ik} \) is non-zero is \( s_B \), and the probability that \( A_{kj} \) is non-zero is \( s_A \). Since the positions of non-zero elements in \( B \) and \( A \) are independently and randomly distributed, the probability that both \( B_{ik} \) and \( A_{kj} \) are non-zero is
\begin{equation}
    \mathbb{P}\left( B_{ik} \neq 0 \text{ and } A_{kj} \neq 0 \right) = s_B s_A.
\end{equation}

Therefore, the probability that \( B_{ik} A_{kj} = 0 \) is
\begin{equation}
    \mathbb{P}\left( B_{ik} A_{kj} = 0 \right) = 1 - s_B s_A.
\end{equation}

Assuming independence across different \( k \), the probability that all terms \( B_{ik} A_{kj} \) are zero is
\begin{equation}
\begin{aligned}
    \mathbb{P}\left( \bigcap_{k=1}^{r} \left\{ B_{ik} A_{kj} = 0 \right\} \right) 
    &= \prod_{k=1}^{r} \mathbb{P}\left( B_{ik} A_{kj} = 0 \right) \\
    &= \left( 1 - s_B s_A \right)^r.
\end{aligned}
\end{equation}

Thus, the probability that \( (BA)_{ij} \) is non-zero is
\begin{equation}
\begin{aligned}
    \mathbb{P}\left( (BA)_{ij} \neq 0 \right) 
    &= 1 - \mathbb{P}\left( (BA)_{ij} = 0 \right) \\
    &= 1 - \left( 1 - s_B s_A \right)^r.
\end{aligned}
\end{equation}

Since there are \( p \times q \) elements in \( BA \), the expected number of non-zero elements is
\begin{equation}
    \mathbb{E}\left[ N_{BA} \right] = p q \left[ 1 - \left( 1 - s_B s_A \right)^r \right],
\end{equation}
where $N_{BA}$ denotes the number of non-zero elements in $BA$.

The expected sparsity of \( BA \) is then
\begin{equation}
    \mathbb{E}\left[s_{BA}\right] = \frac{\mathbb{E}\left[ N_{BA} \right]}{p q} = 1 - \left( 1 - s_B s_A \right)^r.
\end{equation}

The proof of Theorem 3.1 is finished.

\end{proof}

\section{Proof of Theorem 3.2}

\noindent\textbf{Theorem 3.2.} 
\textit{Let \( B \in \mathbb{R}^{p \times r} \) and \( A \in \mathbb{R}^{r \times q} \) be two low rank matrices in LoRA, where the sparsity of \( B \) is \( s_B \) and the sparsity of \( A \) is \( s_A \). Define \( C = BA \), with sparsity \( s_C \). Then, for any \( \delta > 0 \):}
\begin{equation}
\mathbb{P}\left( \left| s_C - \mathbb{E}[s_C] \right| \geq \delta \right) \leq 2 \exp\left( - \dfrac{2 \delta^2 pq}{r(p + q)} \right),
\end{equation}
\textit{where the expected sparsity \( \mathbb{E}[s_C] \) is given by \cref{theorem:sparsity_expectation}
}

\vspace{-0.8em}
\begin{proof}
We aim to apply McDiarmid's inequality to the total number of nonzero entries \( N \) in \( C \).

\textbf{McDiarmid's Inequality} states that if \( X_1, X_2, \dots, X_n \) are independent random variables taking values in a set \( \mathcal{X} \), and \( f: \mathcal{X}^n \rightarrow \mathbb{R} \) satisfies the bounded differences condition: for all \( i \) and all \( x_1, \dots, x_n, x_i' \in \mathcal{X} \),
\[
\left| f(x_1, \dots, x_i, \dots, x_n) - f(x_1, \dots, x_i', \dots, x_n) \right| \leq c_i,
\]
then for all \( \epsilon > 0 \),
\[
\mathbb{P}\left( f(X_1, \dots, X_n) - \mathbb{E}[f] \geq \epsilon \right) \leq \exp\left( -\dfrac{2 \epsilon^2}{\sum_{i=1}^n c_i^2} \right),
\]
and similarly for \( \mathbb{P}\left( \mathbb{E}[f] - f(X_1, \dots, X_n) \geq \epsilon \right) \).

In our context, consider the function \( f \) representing the total number of nonzero entries in \( C \):
\begin{equation}
N = \sum_{i=1}^p \sum_{j=1}^q X_{ij},
\end{equation}
where \( X_{ij} \) is the indicator variable:
\begin{equation}
X_{ij} = \begin{cases}
1, & \text{if } C_{ij} \neq 0, \\
0, & \text{if } C_{ij} = 0.
\end{cases}
\end{equation}

Each \( C_{ij} \) depends on the random variables \( \{ B_{ik}, A_{kj} \}_{k=1}^r \). The variables \( B_{ik} \) and \( A_{kj} \) are independent and affect \( N \) through \( C_{ij} \).

We have the bounded differences:

\textit{Effect of changing \( B_{ik} \):} Changing \( B_{ik} \) can affect all \( C_{ij} \) where \( j = 1, \dots, q \). The maximum change in \( N \) due to changing \( B_{ik} \) is $c_{B_{ik}} = q$.

\textit{Effect of changing \( A_{kj} \):} Changing \( A_{kj} \) can affect all \( C_{ij} \) where \( i = 1, \dots, p \). The maximum change in \( N \) due to changing \( A_{kj} \) is $c_{A_{kj}} = p$.

Therefore, the sum of the squares of the bounded differences is:
\begin{equation}
\sum_{i,k} c_{B_{ik}}^2 + \sum_{k,j} c_{A_{kj}}^2 = pr \cdot q^2 + rq \cdot p^2 = rpq(p + q).
\end{equation}

Applying McDiarmid's inequality, for any \( \epsilon > 0 \):
\begin{equation}
\mathbb{P}\left( N - \mathbb{E}[N] \geq \epsilon \right) \leq \exp\left( -\dfrac{2 \epsilon^2}{rpq(p + q)} \right),
\end{equation}
and similarly for \( \mathbb{P}\left( \mathbb{E}[N] - N \geq \epsilon \right) \). Therefore,
\begin{equation}    
\mathbb{P}\left( \left| N - \mathbb{E}[N] \right| \geq \epsilon \right) \leq 2 \exp\left( -\dfrac{2 \epsilon^2}{rpq(p + q)} \right).
\end{equation}

Since \( s_C = \dfrac{N}{pq} \), we have:
\begin{equation}
\left| s_C - \mathbb{E}[s_C] \right| = \dfrac{\left| N - \mathbb{E}[N] \right|}{pq}.
\end{equation}

Let \( \delta = \dfrac{\epsilon}{pq} \), so \( \epsilon = \delta pq \). Substituting back into the inequality:
\begin{equation}
\begin{aligned}
    \mathbb{P}\left( \left| s_C - \mathbb{E}[s_C] \right| \geq \delta \right) 
    &\leq 2 \exp\left( -\dfrac{2 (\delta pq)^2}{rpq(p + q)} \right) \\
    &= 2 \exp\left( -\dfrac{2 \delta^2 pq}{r(p + q)} \right).
\end{aligned}
\end{equation}

The proof of Theorem 3.2 is finished.

\end{proof}
\section{Proof of Theorem D.1}

\noindent\textbf{Theorem D.1.} 
\textit{Consider matrices \( A \in \mathbb{R}^{r \times q} \) and \( B \in \mathbb{R}^{p \times r} \), where each row of \( B \) and each column of \( A \) exhibit uniform sparsity internally but vary across rows and columns, respectively, with average sparsities \( s_A \) and \( s_B \).
Then, the expected proportion \( \mathbb{E}[ s_C ] \) of nonzero entries in the product matrix \( C = BA \) satisfies:}
\begin{equation}
\mathbb{E}[ s_C ] \leq 1 - (1 - s_A s_B)^r.
\end{equation}

\vspace{-0.8em}
\begin{proof}
Consider any entry $c_{ij}$ of the matrix $C = BA$, which is computed as:
\begin{equation}
c_{ij} = \sum_{k=1}^r b_{ik} a_{kj}.
\end{equation}

To determine the probability that $c_{ij}$ is nonzero, we analyze the sparsity of $b_{ik}$ and $a_{kj}$.

For fixed $i$ and $j$, we define: $s_{B_i}$ is the sparsity of the $i$-th row of $B$; $s_{A_j}$ is the sparsity of the $j$-th column of $A$,.

For $b_{ik}$ and $a_{kj}$, we have:
\begin{equation}
\mathbb{P}(b_{ik} \neq 0) = s_{B_i}, \quad \mathbb{P}(a_{kj} \neq 0) = s_{A_j}.
\end{equation}

Since the positions of nonzero elements within the $i$-th row of $B$ and the $j$-th column of $A$ are independently and uniformly distributed, the events that $b_{ik}$ and $a_{kj}$ are nonzero are independent for each $k$.
Therefore, the probability that both $b_{ik}$ and $a_{kj}$ are nonzero is:
\begin{equation}
\mathbb{P}(b_{ik} \neq 0 \text{ and } a_{kj} \neq 0) = s_{B_i} s_{A_j}.
\end{equation}

Same as the proof for Theorem 3.1, we prove a stronger conclusion by assuming that all elements in $A$ and $B$ are nonnegative. Thus, for $c_{ij} = 0$, it must hold that for all $k = 1, 2, \dots, r$, either $b_{ik} = 0$ or $a_{kj} = 0$. Consequently, the probability that $c_{ij} = 0$ is:
\begin{equation}
\begin{aligned}
    \mathbb{P}(c_{ij} = 0) &= \prod_{k=1}^r \left[ 1 - \mathbb{P}(b_{ik} \neq 0 \text{ and } a_{kj} \neq 0) \right] \\
    &= \left( 1 - s_{B_i} s_{A_j} \right)^r.
\end{aligned}
\end{equation}

Thus, the probability that $c_{ij}$ is nonzero is:
\begin{equation}
\begin{aligned}
    \mathbb{P}c_{ij} \neq 0) &= 1 - \mathbb{P}(c_{ij} = 0) \\
    &= 1 - \left( 1 - s_{B_i} s_{A_j} \right)^r.
\end{aligned}
\end{equation}

Therefore, the expected proportion of nonzero entries in $C$ is:
\begin{equation}
\mathbb{E}[s_C] = \frac{1}{p q} \sum_{i=1}^p \sum_{j=1}^q \left[ 1 - \left( 1 - s_{B_i} s_{A_j} \right)^r \right].
\end{equation}

Note that for $x \in [0, 1]$ and $r \geq 1$, the function $f(x) = (1 - x)^r$ is convex.
According to Jensen's Inequality, for a convex function $f$ and a random variable $X$, we have:
\begin{equation}
\mathbb{E}[f(X)] \geq f(\mathbb{E}[X]).
\end{equation}

In our case, let the random variables be $X_{ij} = s_{B_i} s_{A_j}$, then:
\begin{equation}
\begin{aligned}
    \mathbb{E}[X] &= \frac{1}{p q} \sum_{i=1}^p \sum_{j=1}^q X_{ij} \\
    &= \left( \frac{1}{p} \sum_{i=1}^p s_{B_i} \right) \left( \frac{1}{q} \sum_{j=1}^q s_{A_j} \right) \\
    &= s_B s_A.
\end{aligned}
\end{equation}

Applying Jensen's Inequality, we obtain:
\begin{equation}
\frac{1}{p q} \sum_{i=1}^p \sum_{j=1}^q \left( 1 - s_{B_i} s_{A_j} \right)^r \geq \left( 1 - s_B s_A \right)^r.
\end{equation}

That is:
\begin{equation}
\begin{aligned}
    1 - \mathbb{E}[s_C] &= \frac{1}{p q} \sum_{i=1}^p \sum_{j=1}^q \left( 1 - s_{B_i} s_{A_j} \right)^r \\
    &\geq \left( 1 - s_B s_A \right)^r.
\end{aligned}
\end{equation}

From the inequality above, we have:
\begin{equation}
\mathbb{E}[s_C] \leq 1 - \left( 1 - s_B s_A \right)^r.
\end{equation}

The proof of Theorem 3.3 is finished.

\end{proof}

\section{Proof of LoRASculpt Sparsity Guarantee}

We first demonstrate that incorporating Knowledge-Guided Regularization impacts the sparsity structure of the low-rank LoRA matrices. Specifically, each row of $B$ maintains uniform sparsity, though different rows have varied sparsity levels; similarly, each column of $A$ has consistent sparsity within itself, while sparsity varies across columns. The overall sparsity of the two low-rank matrices remains at $s_B$ and $s_A$ following one-shot pruning.
For the partial derivative of the $(i,j)$-th element of the delta weight $BA$, we have the following expression:
\begin{equation}
\setlength\abovedisplayskip{3pt} \setlength\belowdisplayskip{3pt}
\frac{\partial \mathcal{L}_{\textit{\text{CMR}}}^2}{\partial (BA)_{ij}} = 2 \cdot M_{ij}^2 \cdot \sum_k B_{ik} A_{kj},
\end{equation}

This indicates that the penalty on the $(i,j)$-th position in the delta weight  $BA$  affects the $i$-th row of  $B$  and the $j$-th column of  $A$ . Consequently,  $B$  is constrained by rows, and  $A$  by columns, resulting in varying sparsity across the rows of $B$ and the columns of $A$. Under this condition, the following theorem holds:

\begin{theorem}
\label{theorem:ours_sparsity_expectation}
\vspace{-0.5em}
Consider matrices \( A \in \mathbb{R}^{r \times q} \) and \( B \in \mathbb{R}^{p \times r} \), where each row of \( B \) and each column of \( A \) exhibit uniform sparsity internally but vary across rows and columns, respectively, with average sparsities \( s_A \) and \( s_B \).
Then, the expected proportion \( \mathbb{E}[ s_C ] \) of nonzero entries in the product matrix \( C = BA \) satisfies:
\begin{equation}
\setlength\abovedisplayskip{3pt} \setlength\belowdisplayskip{3pt}
\mathbb{E}[ s_C ] \leq 1 - (1 - s_A s_B)^r.
\end{equation}
\textit{Proof.} See Appendix C.
\hfill\(\Box\)
\vspace{-0.5em}
\end{theorem}

Despite the non-uniform sparsity of matrices $B$ and $A$ across rows and columns, where different rows of $B$ and different columns of $A$ exhibit varied distributions, we can still assume the independence of updates across all elements. This does not hinder the application of McDiarmid’s inequality, thereby allowing us to obtain the previously established error bounds in \cref{theorem:sparsity_range}.
Thus, we have established the sparsity guarantees of \ourmethodslim{}.

\section{Algorithm of LoRASculpt}

The algorithm is outlined in \cref{alg:ours}. Please refer to \cref{sec:Proposed_Method} for more details.

\begin{algorithm}[t]
\setlength{\baselineskip}{1.2\baselineskip}
\caption{LoRASculpt}
\label{alg:ours}
\SetAlgoLined
\SetNoFillComment
\SetArgSty{textnormal}

\small{\KwIn{Training Steps $T$, Warmup Steps $T_{\text{warmup}}$, Training data $\mathcal{D}_{\text{tr}}$, Sparsity Ratio $s_A$, $s_B$ , Number of Layer in LLM and Connector $L_{\textit{LLM}}$, $L_{\textit{Con}}$, Option of whether training Connector with LoRA $\text{Flag}_{\text{\textit{Con}}}$.}}
\small{\KwOut{Final LoRA weights.}}
$\mathcal{L}_{\text{\textit{CMR}}}^{\text{LLM}} \gets 0, \quad \mathcal{L}_{\text{\textit{CMR}}}^{\text{Con}} \gets 0$ \;
$S \gets \psi(W) = \left| 1 / \log\left(\frac{|W|}{\|W\|_2} + \epsilon\right) \right|$ ;       \hfill $\triangleright$ \cref{eq:cal_s} \\
$M \gets \tanh(\omega \odot S)$ ; \hfill $\triangleright$ \cref{eq:cal_m} \\

\For{$t = 1, 2, \dots, T$}{
    Sample a batch $(x^\text{vision}, x^\text{text}, y)$ in $\mathcal{D}_{\text{tr}}$ \;
    \If{$t \geq T_{\text{warmup}}$}{
        \If{$t = T_{\text{warmup}}$}{
            $M_A \gets \operatorname{Mask}(A, s_A)$ ; \\
            $M_B \gets \operatorname{Mask}(B, s_B)$ ; 
            \hfill $\triangleright$ \cref{eq:sparsify} \\
        }
        $A \gets M_A \odot A$ ; \\
        $B \gets M_B \odot B$ ;
        \hfill $\triangleright$ \cref{eq:sparsify_AB} \\
    }
    $h^\text{vision} = \varphi_{\text{\textit{Con}}} \circ \varphi_{\text{\textit{Vis}}} (x^\text{vision}), h^\text{text} = \operatorname{Tokenize}(x^\text{text})$ ;
    
    $\mathcal{L}_{\text{\textit{Task}}} \gets \mathcal{L}_{\text{\textit{CE}}} \Big( \mathit{\Phi} \big[h^\text{vision}, h^\text{text} \big], y \Big)$ \;
    
    \For{$l = 1, 2, \dots, L_{\text{\textit{LLM}}}$}{
        $\mathcal{L}_{\text{\textit{CMR}}}^{\text{LLM}} \gets \mathcal{L}_{\text{\textit{CMR}}}^{\text{LLM}} + \| M_l \odot (B_l A_l) \|_F$ ;
        \hfill $\triangleright$ \cref{eq:reg_cmr} \\
    }
    \If{$\text{Flag}_{\text{\textit{Con}}} = \text{True}$}{
        \For{$\Tilde{l} = 1, 2, \dots, L_{\text{\textit{Con}}}$}{
            $\mathcal{L}_{\text{\textit{CMR}}}^{\text{Con}} \gets \mathcal{L}_{\text{\textit{CMR}}}^{\text{Con}} + \| M_{\Tilde{l}} \odot (B_{\Tilde{l}} A_{\Tilde{l}}) \|_1$ \;
             \hfill $\triangleright$ \cref{eq:loss_conn} \\
        }
    }
    $\mathcal{L} = \mathcal{L}_{\text{\textit{Task}}} + \alpha \cdot \mathcal{L}_{\text{\textit{CMR}}}^{\text{LLM}} + \beta \cdot \mathcal{L}_{\text{\textit{CMR}}}^{\text{Con}}$ ;
    \hfill $\triangleright$ \cref{eq:adapt_connector} \\
    Update low-rank adapters to minimize $\mathcal{L}$ \;
}
\Return Fine-tuned LoRA in $\mathit{\Phi}$ (and $\varphi_{\text{\textit{Con}}}$)
\end{algorithm}
\section{Addition Evaluation Details}

\noindent\textbf{Details of Compared Baselines}

\noindent(a) \lora{} \pub{ICLR'22} \cite{LoRA_ICLR22}: Introduces low-rank adapters to efficiently fine-tune large models.

\noindent(b) \dora{} \pub{ICML'24} \cite{DoRA_ICML24}: Enhances the learning capacity and training stability of LoRA by decomposing weights into magnitude and direction.

\noindent(c) \orthreg{} \pub{ECCV'24} \cite{PEGO_ECCV24}: Adds an orthogonal regularization with a hyperparameter (\textit{i.e.}, 1e-3) to LoRA weights, encouraging fine-tuned features to be orthogonal to pretrained features to preserve model generalization. For fair comparison and due to resource constraints, the component that involves multiple LoRA modules is excluded.

\noindent(d) L2-Regularization \pub{PNAS'17} \cite{EWC_PNAS17}: Apply $L_2$ regularization with a hyperparameter (\textit{i.e.}, 1e-3) to the LoRA weights, guiding the fine-tuned model closer to the pretrained model thus reducing forgetting.

\noindent(e) \dare{} \pub{ICML'24} \cite{DARE_ICML24}: Parameters from the fine-tuned LoRA weights are randomly selected and re-scaled to mitigate knowledge conflict of the target task and other tasks.

\noindent(f) \tailor{} \pub{ICML'24} \cite{ModelTailor_ICML24}: Retains pretrained parameters while selectively replacing a small portion (\textit{i.e.}, 10\%) of fine-tuned parameters, guided by salience and sensitivity analysis.

\vspace{0.3em}
\noindent\textbf{Evaluation Metric.}

To evaluate the performance of MLLMs in general and specialized knowledge, we compute the source performance (denotes by \textit{Source}) and target performance (denotes by \textit{Target}):
\begin{equation}
\setlength\abovedisplayskip{3pt} \setlength\belowdisplayskip{3pt}
\textit{Source} = \frac{1}{|\mathcal{D}|} \sum_i^{|\mathcal{D}|} \operatorname{Score}(\mathcal{D}_i), \;\;\; \textit{Target} = \operatorname{Score}(\mathcal{T}).
\label{eq:metrics}
\end{equation}
where $\operatorname{Score}(\cdot)$ denotes the evaluation metric for different datasets, which is set to Accuracy and CIDEr for VQA and Captioning tasks, respectively. 
Here, $\mathcal{D}=\{\mathcal{D}_i\}_{i=1}^{|\mathcal{D}|}$  represents the datasets used to evaluate general knowledge, and $\mathcal{T}$ denotes the downstream task dataset. We use the average score of \textit{Source} and \textit{Target}, denoted as \textit{Avg} to measure the overall capability of the MLLM.

\section{Ablation Study of \texorpdfstring{$\beta$}{beta}}

$\beta$ controls the sparsity strength for MLLM connector in \cref{eq:adapt_connector}. Since the connector plays a crucial role in modality alignment, adopting a high sparsity level could lead to performance degradation on downstream tasks (denoted by \textit{Target} in \cref{tab:ablation_beta}). Selecting an appropriate $\beta$ to sparsify the connector can achieve a balance between \textit{Source} and \textit{Target}.

\begin{table}[h]\small
\centering
{
\renewcommand\arraystretch{1.1}
\begin{tabular}{c||ccccc}
\hline \thickhline
\rowcolor{lightgray}
$\beta$ & $10^{-2}$ & $10^{-3}$ & $10^{-4}$ & $10^{-5}$ & $10^{-6}$  \\
\hline\hline
\textit{Source} & 60.19 & 58.58 & 59.73 & 59.55 & 59.13  \\
\textit{Target} & 80.10 & 79.99 & 84.02 & 85.34 & 85.01  \\
\rowcolor[HTML]{D7F6FF}
\textit{Avg}    & 70.15 & 69.29 & 71.87 & \textbf{72.45} & 72.07  \\
\hline
\end{tabular}}
\vspace{-5pt}
\captionsetup{font=small}
\caption{
\textbf{Ablation Study of $\beta$}, which represents the intensity of sparsity applied to the MLLM connector. When set to $10^{-5}$, the optimal \textit{Avg} is achieved. 
}
\label{tab:ablation_beta}
\vspace{-5pt}
\end{table}

{
    \small
    \bibliographystyle{ieeenat_fullname}
    \bibliography{main}
}

\end{document}